\documentclass[pdflatex,sn-mathphys-num]{sn-jnl}%

\usepackage{graphicx}%
\usepackage{multirow}%
\usepackage{amsmath,amssymb,amsfonts}%
\usepackage{amsthm}%
\usepackage{mathrsfs}%
\usepackage[title]{appendix}%
\usepackage{xcolor}%
\usepackage{textcomp}%
\usepackage{manyfoot}%
\usepackage{booktabs}%
\usepackage{algorithm}%
\usepackage{algorithmicx}%
\usepackage{algpseudocode}%
\usepackage{listings}%
\usepackage{bm}
\usepackage{subfig}

\theoremstyle{thmstyleone}%

\theoremstyle{thmstyletwo}%

\theoremstyle{thmstylethree}%

\begin{document}
\title[Article Title]{End-to-End Self-Supervised RGB-T Tracking without Modality Misleading}

\author[1,2,3]{\fnm{Shenglan} \sur{Li}}\email{shenglanli@cumt.edu.cn}

\author*[1,2]{\fnm{Rui} \sur{Yao}}\email{ruiyao@cumt.edu.cn}

\author[1,2]{\fnm{Kunyang} \sur{Sun}}\email{kunyang\_sun@cumt.edu.cn}

\author[3]{\fnm{Hong} \sur{Jia}}\email{hong.jia@auckland.ac.nz}

\author[1,2]{\fnm{Yong} \sur{Zhou}}\email{yongzhou@cumt.edu.cn}

\author[4]{\fnm{Javen Qinfeng} \sur{Shi}}\email{javen.shi@adelaide.edu.au}

\author*[3]{\fnm{Xinyu} \sur{Zhang}}\email{xinyu.zhang@auckland.ac.nz}

\affil[1]{\orgdiv{School of Computer Science and Technology / School of Artificial Intelligence, China University of Mining and Technology, China}}
\affil[2]{\orgdiv{Mine Digitization Engineering Research Center of the Ministry of Education, China}}
\affil[3]{\orgdiv{University of Auckland, Auckland, New Zealand}}
\affil[4]{\orgdiv{Australian Institute for Machine Learning, Adelaide University, Australia}}

\footnotetext{The contribution of Rui Yao was supported by the National Natural Science Foundation of China (Grant No. 62676414 and No. 62172417) and the contribution of Shenglan Li was supported by the China Scholarship Council (Grant No. 202506420043).}
\abstract{RGB-T object tracking leverages the complementary characteristics of visible and thermal infrared modalities to improve robustness under adverse conditions. Existing supervised methods typically rely on costly modality-aligned bounding box annotations, while most self-supervised approaches follow a two-stage pseudo-labeling paradigm, making tracker training sensitive to pseudo-label quality and 
preventing joint end-to-end optimization. 
In this paper, we propose ESMTrack, a fully end-to-end self-supervised RGB-T tracking framework 
without offline pseudo-label generation or dense frame-level bounding box annotations.
Given only the standard initial-frame annotation used in visual tracking, ESMTrack learns discriminative and temporally consistent representations through two complementary objectives: a grounding triplet loss on annotated initial frames and a cross-frame temporal triplet loss on unlabeled search frames, with reliable samples selected by forward-backward consistency. 
To address modality dominance bias, ESMTrack employs a three-branch architecture consisting of a fusion branch and two unimodal branches for RGB and thermal inputs. 
We quantify modality contributions using the Average Peak-to-Correlation Energy by measuring response discrepancies between the fusion and unimodal branches. 
The resulting reliability estimates guide a training-time modality decoupling mechanism that suppresses dominant-modality shortcuts and adaptively weights cross-modal contrastive learning for task-level alignment. 
Extensive experiments on five RGB-T tracking benchmarks show that ESMTrack achieves competitive state-of-the-art performance, strong cross-dataset generalization, and real-time inference speed. The source code is available at https://github.com/LiShenglana/ESMTrack.
}


\maketitle

\section{Introduction}\label{sec1}
RGB-T object tracking~\cite{AttriRGBTTracking,IPL} aims to localize a target object over time by exploiting the complementary characteristics of visible and thermal infrared modalities. Compared with RGB-only tracking, RGB-T tracking can offer improved robustness under challenging conditions such as low illumination, occlusion, background clutter, and adverse weather. 
Despite recent progress, most high-performing RGB-T trackers are trained in a fully supervised manner and rely on densely annotated, modality-aligned bounding boxes. Such annotations are expensive to acquire, particularly for large-scale RGB-T videos, where visible and thermal streams must be accurately synchronized and labeled frame by frame. This annotation burden limits the scalability of supervised RGB-T tracking and motivates the development of self-supervised RGB-T tracking methods that can learn from videos with only the standard initial-frame annotation used in visual tracking.

\begin{figure}[t]
  \centering
  \includegraphics[width=\linewidth]{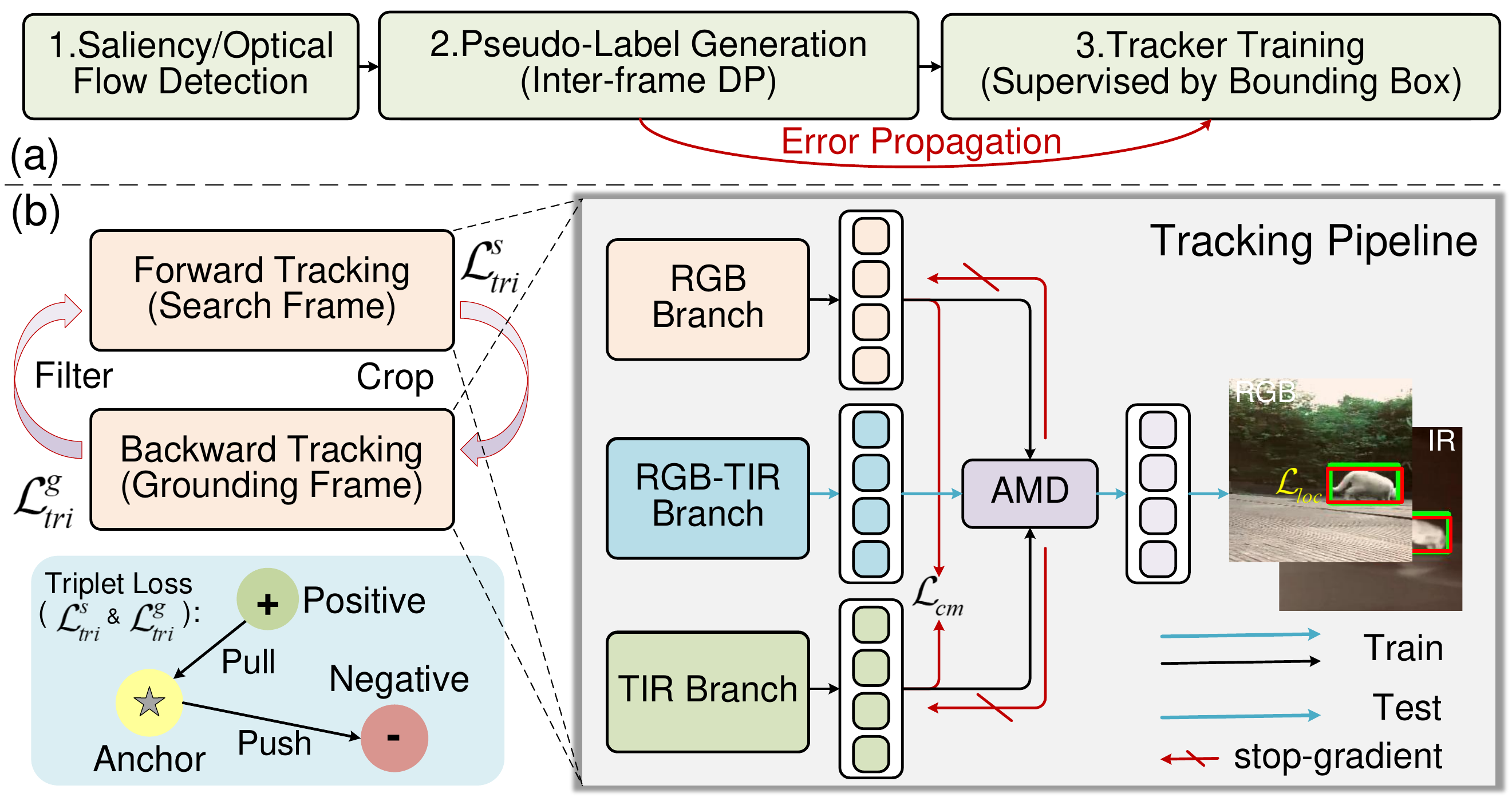}
  \caption{(a) Conventional Two-stage Self-Supervised Paradigm. (b) Our End-to-End Box Free Framework where forward tracking and backward tracking adopt the same tracking pipeline}
  \label{motivation}
\end{figure}

Existing self-supervised RGB tracking methods~\cite{UDT,cyclesiam,PUL,SSTrack} have demonstrated the feasibility of learning tracking representations without dense frame-level annotations. 
However, extending them from single modality to RGB-T scenarios introduces new challenges. 
Most existing self-supervised RGB-T tracking methods~\cite{S2OTFormer, GDSTrack} follow a two-stage paradigm of pseudo-label generation and tracker training, as illustrated in Figure~\ref{motivation}(a). For example, GDSTrack~\cite{GDSTrack} follows the pseudo-label generation paradigm of USOT~\cite{USOT}, utilizing dynamic graphs for bi-modal feature fusion and temporal diffusion to alleviate the adverse effects of inaccurate pseudo-labels. However, such methods suffer from two fundamental flaws. First, tracker training is strongly constrained by the quality of pseudo-labels. Since pseudo-labels are typically generated via saliency detection or optical flow without direct ground-truth guidance, they often exhibit significant spatial deviations from the true target location, introducing systematic noise into the training process that cannot be corrected through subsequent optimization. Second, the decoupled two-stage pipeline trains the pseudo-label generator and the tracker are trained independently, which increases the training complexity and prevents the whole tracking framework from being learned in an end-to-end manner.
As a result, errors introduced during pseudo-label generation may propagate into tracker learning and are difficult to correct during training.
Beyond the challenge of pseudo-label dependency, achieving mutual learning between modalities without mutual interference remains particularly difficult in the absence of ground-truth supervision. Most existing trackers~\cite{TBSI, AINet, TUMFNet} exploit modal complementarity through soft fusion based on feature activation intensity. However, this approach carries a critical defect: feature activation intensity is not necessarily aligned with the dynamic robustness of the tracking task. When RGB features are strongly activated by occlusions or background distractors, high response values may mislead the modal weight distribution, causing unreliable modalities to corrupt the learning process of reliable ones during self-supervised training.

To address these issues, we propose ESMTrack, an end-to-end self-supervised RGB-T tracking framework, as shown in Figure~\ref{motivation}(b). 
Instead of generating offline pseudo-labels as direct supervision, ESMTrack learns from 
a forward-backward tracking consistency through two complementary triplet constraints. 
For unlabelled search frames, we introduce a \textbf{cross-frame temporal triplet loss} to encourage the predicted target representation to remain consistent with the object representation in the annotated initial frame.
Since predictions on search frames may be unreliable, we further perform backward tracking from the predicted search-frame target back to the initial frame, and use the resulting backward IoU as a reliability criterion to filter training samples.
For the annotated initial frame, we introduce a \textbf{grounding triplet loss} that pulls the predicted target region toward the ground-truth object region while pushing it away from spatially shifted hard negatives.
Together, these two losses jointly provide complementary supervision for spatial discriminability and temporal semantic consistency, enabling end-to-end self-supervised training without offline pseudo-label generation or frame-level bounding-box annotations.

Furthermore, to alleviate modality dominance bias during self-supervised cross-modal learning, we propose an \textbf{Adaptive Modality Decoupling} (AMD) module.
Rather than estimating modality importance from raw feature activation intensity, AMD uses task-level response quality to measure the contribution of each modality. 
Specifically, based on the outputs of the fusion branch and two unimodal branches, AMD estimates the current contribution of each modality to the fused tracking representation.
During training, AMD partially suppresses the dominant modality contribution from the fused representation, forcing the fusion branch to mine complementary information from the less dominant modality, which reduces shortcut learning and improves robustness to modality degradation. 
At inference time, the tracker uses only the learned fusion branch, without additional branches or suppression operations.

In summary, our main contributions are as follows:
\begin{itemize}
    \item We propose ESMTrack, an end-to-end self-supervised RGB-T tracking framework that exploits forward-backward tracking consistency to learn from RGB-T videos using only the standard initial-frame annotation, without offline pseudo-label generation or dense frame-level bounding-box annotations.

    \item We introduce two complementary triplet losses, i.e., a cross-frame temporal triplet loss on unlabeled search frames and a grounding triplet loss on annotated initial frames, jointly enhancing spatial discriminability and temporal target consistency while avoiding direct supervision from noisy pseudo-labels.

    \item We propose the Adaptive Modality Decoupling (AMD) module that estimates modality contribution from task-level response quality rather than feature activation intensity, acting as a training-time modality decoupling mechanism that suppresses dominant-modality shortcuts and encourages cross-modal information.
    
    \item Extensive experiments on five RGB-T tracking benchmarks demonstrate that ESMTrack achieves competitive state-of-the-art performance among self-supervised RGB-T trackers, with strong generalization and fast inference speed.
        
\end{itemize}

\section{Related works}\label{sec2}
\subsection{RGB-T Tracking}
RGBT tracking~\cite{PURA,CAFormer,after,AINet} aims to leverage the complementary characteristics of RGB and infrared modalities to achieve robust all-weather object tracking. Supervised RGB-T methods have achieved remarkable progress from multiple perspectives. From the perspective of feature fusion, CADTrack~\cite{CADTrack} integrates Mamba-based feature interaction, MoE-driven contextual aggregation, and deformable temporal alignment to address modality discrepancies, while MambaVT~\cite{MambaVT} introduces a pure Mamba-based framework that effectively models spatio-temporal dependencies in RGB-T sequences. From the perspective of modal quality and reliability, QSTNet~\cite{QSTNet} introduces a quality-aware spatio-temporal transformer that suppresses low-quality tokens by estimating token reliability from search-template correlations. From the perspective of temporal modeling, VCT~\cite{VCT} proposes a video-level dual-stream framework with cross-modal temporal prompt navigation and modality-specific mixture of adapters to capture complementary spatiotemporal information. From the perspective of prompt learning, ViPT~\cite{ViPT} introduces prompt learning into multimodal tracking for flexible cross-modal interaction, and SeqTrackV2~\cite{SeqTrackv2} employs task-specific prompts to unify tracking across thermal, depth, and event modalities. SFPT~\cite{SFPT} further introduces language-derived semantic anchors and adaptive cross-modal fusion to suppress perturbation-induced noise under adversarial conditions.

Despite their impressive performance, these supervised methods share a common limitation in modality fusion: they typically rely on feature activation intensity as a proxy for modality reliability. Although QSTNet~\cite{QSTNet} attempts to model token-level quality via search-template correlation, it remains anchored to feature response magnitude rather than task-level tracking contribution, and thus remains susceptible to modality misleading when a strongly activated but unreliable modality dominates the fusion process and suppresses complementary information from the other. Furthermore, all these methods depend heavily on large-scale annotated data, whose high acquisition costs restrict practical deployment. In contrast, our method adopts a self-supervised paradigm and explicitly addresses modality misleading through task-level response quality estimation, dynamically suppressing the dominant modality to force complementary cross-modal representation learning.
\subsection{Self-supervised Tracking}
Self-supervised tracking addresses the challenge of requiring large-scale manual annotations for tracking tasks. Most self-supervised tracking methods rely primarily on the RGB modality alone. The UDT~\cite{UDT} exploits the discriminative power of correlation filtering for tracking and employs cycle consistency as a self-supervised signal. Existing deep trackers can be trained using synthesized data in routine ways, without requiring human annotation. PUL~\cite{PUL} employs EdgeBox~\cite{EdgeBoxes} to generate high-quality proposals and leverages contrastive learning to facilitate temporal training sample collection for model supervision. S2Siamfc~\cite{S2Siamfc} exploit the fact that an image and any cropped region of it naturally form a pair for self-training. To replace naive cropping methods such as center cropping, USOT~\cite{USOT} propose a more accurate pseudo-label generation method based on optical flow and dynamic programming techniques. Building on this, \cite{Diff-Tracker} learn a prompt representation of an object using a pre-trained diffusion model. SSTrack~\cite{SSTrack} introduces a decoupled spatio-temporal consistency framework based on instance contrastive loss, enabling robust object learning through global spatial and local temporal modeling. S2OTFormer~\cite{S2OTFormer} proposes a self-supervised RGBT tracking framework that mitigates inaccurate pseudo-label interference through multi-modal feature fusion and motion-aware consistency learning. GDSTrack~\cite{GDSTrack} utilizes dynamic graph fusion and temporal diffusion to mitigate pseudo-label noise triggered by similar object distractors for RGB-T tracking. Yet, exploring intrinsic self-supervised signals from both modalities to achieve pseudo-label-free, end-to-end self-supervised RGB-T tracking warrants further investigation.
\section{Methodology}\label{sec3}
\subsection{Framework Overview}\label{subsec1}
\begin{figure*}[t]
  \centering
  \includegraphics[width=\linewidth]{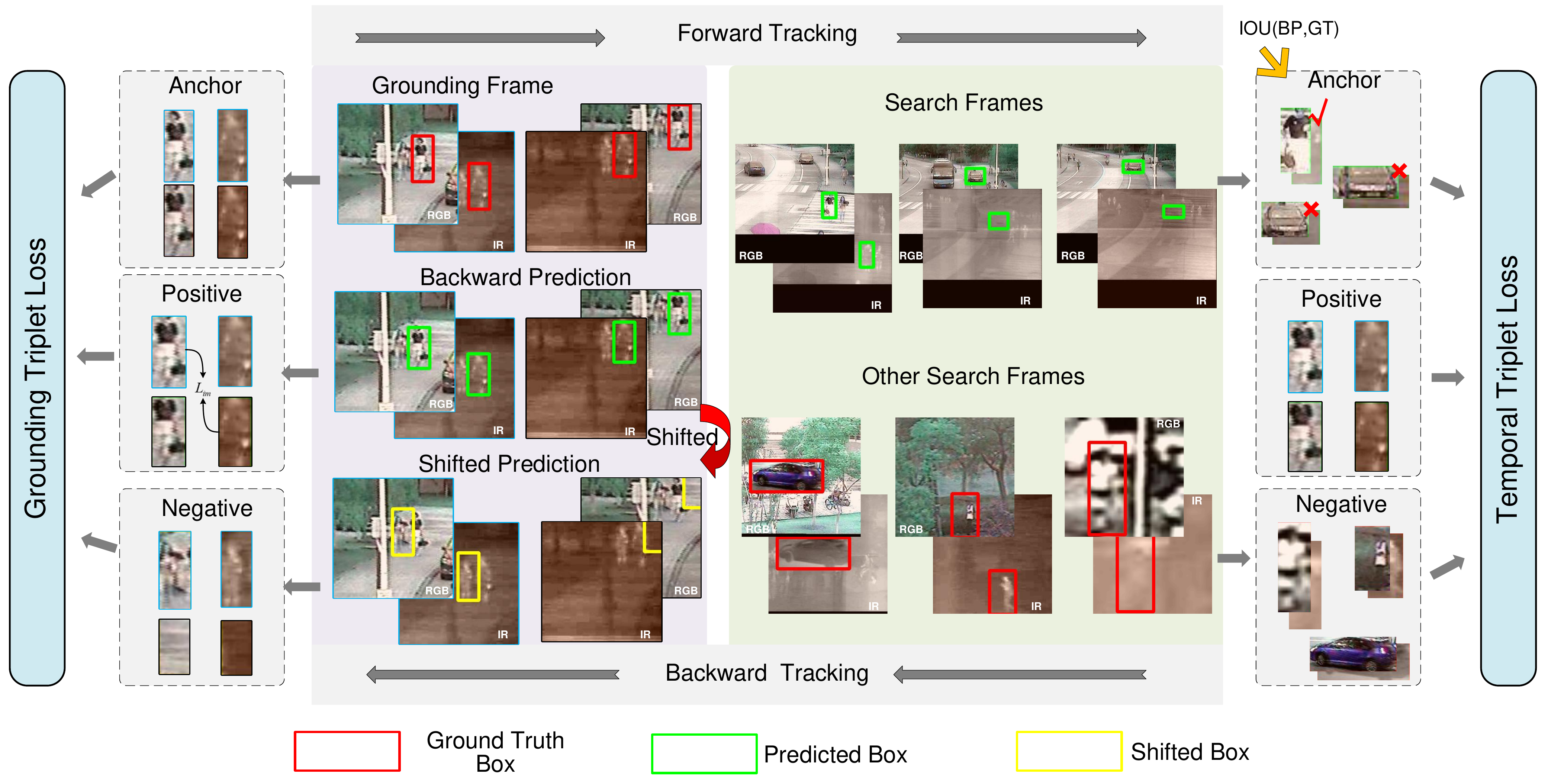}
  \caption{The pipeline of ESMTrack, which learns forward-backward tracking consistency through two complementary triplet constraints: a cross-frame temporal triplet loss for unlabelled search frames with backward IoU filtering to ensure temporal semantic consistency, and a grounding triplet loss for the annotated initial frame to enhance spatial discriminability
}
  \label{pipeline}
\end{figure*}

Our proposed framework, ESMTrack, is illustrated in Figure~\ref{pipeline}. ESMTrack learns tracking consistency through a forward-backward paradigm. Two complementary triplet losses are introduced: a cross-frame temporal triplet loss for unlabelled search frames to enforce temporal semantic consistency, and a grounding triplet loss for the annotated initial frame to ensure spatial discriminability.
For both forward and backward tracking, as shown in Figure~\ref{pipeline_amd}, the framework consists of three branches: a modality fusion branch that achieves preliminary RGB-thermal fusion via prompt-tuning on a frozen pre-trained foundation model, along with two unimodal branches that extract pure RGB and thermal features. Based on the outputs of these three branches, the Adaptive Modal Discrimination (AMD) module dynamically evaluates the contribution of each modality and suppresses redundant or misleading cross-modal features, driving the fusion branch to actively mine complementary information while reducing shortcut dependencies and enhancing the model's robustness to modality degradation.
\subsection{Multi-Modal Input and Feature Extraction}
Self-supervised tracking aims to train a tracker on unlabeled video sequences to accurately localize the object in subsequent frames, given only the bounding box annotation in the first frame. 
ESMTrack takes three types of inputs. The first is a pair of visible and infrared 
template frames, $\boldsymbol{z}^v \in \mathbb{R}^{3 \times H_z \times W_z}$ and 
$\boldsymbol{z}^i \in \mathbb{R}^{1 \times H_z \times W_z}$, cropped tightly around 
the target bounding box in the initial frame and used as the reference in forward 
tracking. The second is a pair of visible and infrared grounding frames, 
$\boldsymbol{g}^v \in \mathbb{R}^{3 \times H \times W}$ and 
$\boldsymbol{g}^i \in \mathbb{R}^{1 \times H \times W}$, cropped from the same 
initial frame with a larger spatial context and used as the search 
region in backward tracking. The third is $T$ pairs of search frames, 
$\{\boldsymbol{x}_t^v\}_{t=1}^T \in \mathbb{R}^{T \times 3 \times H \times W}$ and 
$\{\boldsymbol{x}_t^i\}_{t=1}^T \in \mathbb{R}^{T \times 1 \times H \times W}$, 
sampled from subsequent frames of the sequence for forward tracking.
All visible and infrared inputs are tokenized by two independent patch embedding 
layers into $D$-dimensional token sequences, yielding 
$\boldsymbol{f}_{z}^v, \boldsymbol{f}_{z}^i \in \mathbb{R}^{N_z \times D}$, 
$\boldsymbol{f}_{g}^v, \boldsymbol{f}_{g}^i \in \mathbb{R}^{N_x \times D}$, and 
$\boldsymbol{f}_{x}^v, \boldsymbol{f}_{x}^i \in \mathbb{R}^{N_x \times D}$, 
where $N_z$, $N_x$ denote the patch counts for each frame type.

\begin{figure*}[t]
  \centering
  \includegraphics[width=\linewidth]{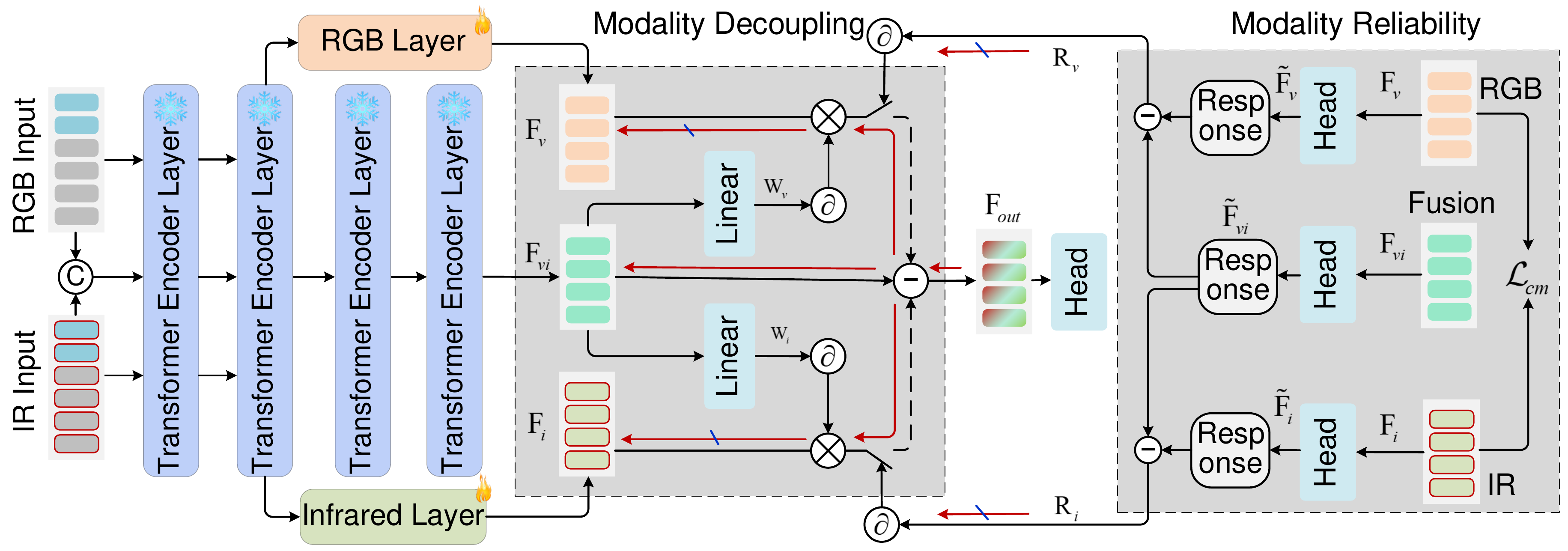}
  \caption{Forward tracking framework, where the fusion branch computes modality importance with each unimodal branch, and the Adaptive Modality Decoupling module suppresses the relatively more dominant modality to enhance model robustness}
  \label{pipeline_amd}
\end{figure*}

The backbone consists of $L$ transformer layers and processes three parallel branches: a \textbf{Fusion branch} that receives the concatenated template-search tokens, an \textbf{RGB branch}, and an \textbf{IR branch}. As illustrated in Figure~\ref{pipeline_amd}, the first $K$ layers share weights across all three branches. Within each of these shared layers, a lightweight cross-modal prompt module receives the concatenated RGB and IR spatial features and produces modality-interaction tokens that are injected back into the fusion branch via a gated IR injection operator, enabling progressive cross-modal information exchange without additional backbone parameters. In the final $L-K$ layers, the three branches diverge with independent parameters. The fusion branch processes the fused representation and outputs $\mathbf{F}_{vi} \in \mathbb{R}^{N_x \times D}$. The RGB branch and IR branch each maintain a separate copy of the transformer layers, producing unimodal search-region features $\mathbf{F}_{v} \in \mathbb{R}^{N_x \times D}$ and $\mathbf{F}_{i} \in \mathbb{R}^{N_x \times D}$, respectively. These three outputs are subsequently fed into the AMD module to produce the final object representation. The backward tracking pass adopts the identical 
architecture, with the sole difference that the cross-modal contrastive loss 
$\mathcal{L}_{\text{cm}}$ is omitted.
\subsection{Adaptive Modality Decoupling with Response-Guided Gating}\label{subsec2}
Existing RGB-T fusion methods rely on feature activation intensity for modality reliability assessment, which can mislead weight distribution under occlusion or distractor interference and cause unreliable modality dominance to pollute cross-modal representation learning in self-supervised settings. To address this limitation, we propose an Adaptive Modality Decoupling mechanism with Response-Guided Gating that replaces feature activation intensity with task-level tracking confidence as the criterion for modality importance estimation. Specifically, the procedure consists of two stages: APCE-Based Modality Reliability Estimation and Gated Adaptive Modality Decoupling. By dynamically identifying and actively suppressing the currently dominant modality during training, the fusion branch is forced to continuously mine complementary information from the less dominant modality, thereby preventing shortcut learning and improving robustness against modality degradation at test time. 

\textbf{APCE-Based Modality Reliability Estimation.} For each frame, we quantify the contribution of each modality by simulating single-modality inference. All three feature sets are passed through the tracking head to produce response maps, from which we compute the Average Peak-to-Correlation Energy (APCE)~\cite{LMCF}:
\begin{align}
\text{APCE}(\mathbf{S}) = \frac{(S_{max} - S_{min})^2}{\frac{1}{HW}\sum_{i}(S_i - S_{min})^2}.
\end{align}
A higher APCE indicates a more peaked, confident response map. Taking RGB as an example, we measure the response degradation by comparing the fusion branch against the TIR unimodal branch:

\begin{align}
\Delta_{v} = \text{APCE}(\mathbf{\tilde{F}}_{vi}) - \text{APCE}(\mathbf{\tilde{F}}_{i}),
\end{align}
where $\mathbf{\tilde{F}}_{vi}$ denotes the response map of the fusion branch and $\mathbf{\tilde{F}}_{i}$ denotes that of the TIR unimodal branch. A large $\Delta_{v}$ indicates that the fusion branch significantly outperforms the TIR-only branch, implying that the RGB modality makes a strong contribution to the current fusion representation. The modality importance score is then defined as:
\begin{align}
R_{v} = \sigma\left(\frac{\Delta_{v}}{T}\right),
\end{align}
where $\sigma(\cdot)$ is the sigmoid function and T is a temperature coefficient (set as 0.5). The TIR importance score $R_{i}$ is computed symmetrically. They are subsequently used to gate the subtraction of unimodal features from the fusion representation.

\textbf{Gated Adaptive Modality Decoupling.} Given the reliability scores $R_{v}$ and $R_{i}$, we introduce a Gated Cross-Modal Fusion module that actively removes the single-modal contribution of whichever modality is currently reliable from the fused representation:
\begin{align}
\mathbf{F}_{out} = \mathbf{F}_{vi} - G_{v} \cdot \mathbf{F}_{v} - G_{i} \cdot \mathbf{F}_{i}
\end{align}
The gate weights are defined as:
\begin{equation}
G_{\mathrm{v}} = 
\begin{cases} 
R_{\mathrm{v}} \cdot \sigma\left(\mathbf{W}_{v}(\mathbf{F}_{\mathrm{vi}})\right), & \text{if } R_{\mathrm{v}} > \tau \\
0, & \text{otherwise},
\end{cases}
\end{equation}
and $G_{i}$ is defined analogously, where $\tau$ (set as 0.2) is a hard threshold and $\mathbf{W}_{v}, \mathbf{W}_{i} \in \mathbb{R}^{C \times C}$ are learnable linear projections. To prevent the suppression gradient from interfering with the unimodal encoders, $\mathbf{F}_{v}$ and $\mathbf{F}_{i}$ are detached from the computation graph before gating, ensuring that each encoder is trained solely through the main tracking loss rather than through the AMD suppression path. The design rationale is as follows: when a modality is reliable ($w > \tau$), its single-modal features are subtracted from the fused representation in proportion to the gate strength, effectively applying feature-level dropout to that modality. Then the model can no longer exploit the dominant modality freely and is compelled to extract complementary information from the other modality. Importantly, the suppression magnitude is continuously modulated by sigmoid gate, which is bounded in $(0,1)$ and learned end-to-end. This continuous formulation ensures that even when one modality persistently dominates across many training samples, the suppression remains partial rather than complete, preserving residual gradient signal and preventing the fused pathway from becoming entirely blind to the dominant modality. Conversely, when a modality is of low quality ($R \leq \tau$), the gate is set to zero and the fused representation is left unchanged, avoiding further suppression of an already weak signal. Crucially, when both modalities are simultaneously degraded ($R_{\mathrm{v}} \leq \tau$ and $R_{\mathrm{i}} \leq \tau$), both gates collapse to zero and the module reduces to training on the pure fused branch without any suppression. This fallback prevents the gating from amplifying unreliable APCE estimates under joint degradation, where neither modality provides a trustworthy signal to guide suppression. By continuously imposing this cross-modal learning pressure during training, the model develops robust responses to arbitrary single-modality degradation at inference.
\subsection{Triplet Loss for Grounding and Search Frames}
To address the systematic noise introduced by pseudo-label dependency and enable end-to-end self-supervised training without relying on an offline pseudo-label generator, we design two complementary triplet losses tailored to the distinct supervision conditions of grounding and search frames. For unlabelled search frames, where ground-truth annotations are unavailable, the Cross-frame Temporal Triplet Loss maintains semantic coherence between the predicted target representation and the initial-frame object representation, with training samples filtered by backward tracking IoU, while pushing it away from cross-sequence negatives sampled from other object identities. For the annotated grounding frame, the Grounding Triplet Loss provides accurate spatial supervision by pulling the predicted target toward the ground-truth region while repelling spatially shifted box regions as hard negatives, anchoring the model with reliable supervision to counteract the drift accumulated during search-frame self-supervision.

\textbf{Cross-frame Temporal Triplet Loss.} For search frames, ground-truth annotations are not available during training; instead, the model relies on its own predicted boxes. To nonetheless impose a cross-frame consistency constraint, we introduce a complementary triplet loss that bridges search-frame predictions with grounding-frame ground truth. The anchor $\mathbf{f}_{a}^{s}$ is the pooled feature at the predicted box location in the search frame. The positive $\mathbf{f}_{p}^{s}$ is the pooled feature at the ground-truth box location in the temporally corresponding grounding frame, capturing the canonical object appearance. The negative $\mathbf{f}_{n}^{s}$ is constructed by rolling the batch of positive features by one position, so that each sample's negative is a different sample's object identity, which can guarantee a semantically distinct negative without requiring annotations.

To ensure the anchor features are reliable, the Intersection over Union (IoU) of the backward tracking serves as a filtering criterion to assess the accuracy of forward tracking pseudo-labels. This addresses the limitation in previous cycle-consistency training where gradients failed to guide the forward tracking process. Specifically, we compute $\text{IoU}_{bwd}$ by measuring the overlap between the model's prediction on the grounding frame and the grounding frame's ground-truth box. Only samples with $\text{IoU}_{bwd} > \tau$ are included, filtering out cases where the tracker has already drifted to an incorrect object. The temporal triplet loss is computed over all pairwise combinations:
\begin{align}
\mathcal{L}_{tri}^{s} = \max\bigl(D(\mathbf{f}_a^{s}, \mathbf{f}_p^{s}) - D(\mathbf{f}_a^{s}, \mathbf{f}_n^{s}) + \alpha; 0\bigr).
\end{align}
This loss encourages the search-frame feature at the predicted location to be more similar to the correct object's grounding representation than to any other object's representation, effectively tightening cross-frame object identity consistency.

\textbf{Grounding Frame Triplet Loss.} To guide the model's localization accuracy on grounding frames, we introduce a triplet loss that operates directly on predicted bounding box regions. For each grounding frame, we construct three region-level feature representations via masked average pooling over the encoder output tokens. 

Specifically, the anchor $\mathbf{f}_{a}^g$ is obtained by pooling features within the ground-truth bounding box region $\mathbf{b}_{gt}$, the positive $\mathbf{f}_{p}^g$ is obtained by pooling within the predicted bounding box region $\mathbf{b}_{pre}$, and the hard negative $\mathbf{f}_{n}^g$ is obtained by pooling within a spatially perturbed box region $\mathbf{b}_{gt}$. The perturbation is constructed by applying a random displacement to the ground-truth box center, with the shift magnitude drawn from a mixed distribution where 30\% of samples receive 1.5× the base shift magnitude $\delta = 0.08$, and the displacement is clamped to within 70\% of the object's width and height to ensure spatial adjacency. The triplet loss is then defined as:
\begin{align}
\mathcal{L}_{tri}^{g} = \max\bigl(D(\mathbf{f}_{a}^g, \mathbf{f}_{p}^g) - D(\mathbf{f}_{a}^g, \mathbf{f}_{n}^g) + \alpha; 0\bigr),
\end{align}
where $D(\cdot,\cdot) = 1 - \cos(\cdot,\cdot)$ denotes the cosine distance. The anchor and negative features are detached from the gradient computation, so the loss updates only the predicted-region branch, pulling the tracker's predicted localization feature toward the ground-truth object region while pushing it away from the hard nearby distractor.
\subsection{Training and Inference}
\textbf{Training Objective.} The overall training loss combines five components:
\begin{align}
\mathcal{L} = \mathcal{L}_{loc} + \beta\mathcal{L}_{cm} + \mathcal{L}_{cont} + \mathcal{L}_{rgb\text{-}tir} + \lambda\mathcal{L}_{tri}^{g} + \gamma\mathcal{L}_{tri}^{s}.
\label{ParameterWeight}
\end{align}
Localization loss $\mathcal{L}_{loc}$ is computed exclusively on grounding frames, which carry ground-truth box annotations, providing IoU regression supervision for accurate target localization. It consists of three terms: a GIoU loss (weight 2.0), an $\ell_1$ bounding box regression loss (weight 5.0), and a Focal loss on the predicted Gaussian heatmap:
\begin{align}
\mathcal{L}_{loc} = \lambda_{giou}\mathcal{L}_{giou} + \lambda_{l_1}\mathcal{L}_{1} + \mathcal{L}_{focal}.
\end{align}
Cross-modal alignment loss $\mathcal{L}_{cm}$ is a bidirectional InfoNCE loss $\mathcal{L}_{IN}$ between the RGB and TIR single-modal feature streams, computed on search frames:
\begin{equation}
\begin{aligned}
\mathcal{L}_{cm} = & R_{\mathrm{v}} \cdot \mathcal{L}_{IN}(\mathbf{F}_{\mathrm{v}} \to \mathbf{F}_{\mathrm{i}}) + R_{\mathrm{i}} \cdot \mathcal{L}_{IN}(\mathbf{F}_{\mathrm{i}} \to \mathbf{F}_{\mathrm{v}}).
\end{aligned}
\end{equation}
$\mathcal{L}_{cm}$ is weighted by importance to constrain representation learning at the task level, thereby preventing inter-modal misleading.
Following~\cite{SSTrack}, multi-view instance contrastive loss $\mathcal{L}_{cont}$ is an NT-Xent loss applied between the fused features of two independently augmented views of the same grounding frame within each training sample, encouraging the fused representation to be invariant to appearance variation introduced by random cropping and photometric augmentation:
\begin{align}
\mathcal{L}_{cont} = \mathcal{L}_{NT}(\mathbf{F}_{vi}^{g_1},\mathbf{F}_{vi}^{g_2}).
\end{align}
The RGB-TIR cross-view contrastive loss $\mathcal{L}_{rgb\text{-}tir}$ is an NT-Xent loss applied cross-modally between region-level feature representations, obtained via masked average pooling over the encoder output tokens of the augmented views $\mathbf{f}_{v/i}^{g_1}, \mathbf{f}_{v/i}^{g_2}$ of the grounding frame, pulling together representations of the same object observed in different modalities:
\begin{align}
\mathcal{L}_{rgb\text{-}tir} = \mathcal{L}_{NT}(\mathbf{f}_{v}^{g_1},\mathbf{f}_{i}^{g_2}) + \mathcal{L}_{NT}(\mathbf{f}_{v}^{g_2}, \mathbf{f}_{i}^{g_1}).
\end{align}

\textbf{Inference.} At inference time, the tracker operates in a strictly online, single-pass manner without any additional post-processing. Given the initial object bounding box in the first frame, a template crop is extracted and encoded once. For each subsequent frame, a search region is cropped around the last predicted location, concatenated channel-wise with the corresponding thermal patch, and fed into the backbone together with the template tokens.
During inference, only the fusion branch is retained, with the unimodal branches and suppression operations discarded, imposing no additional computational overhead.
The encoder output $\mathbf{F}_{vi}$ is used directly, as it has already internalized complementary cross-modal representations through the modality decoupling constraints imposed during training. The fused features are then passed to the head to predict the object bounding box for the current frame, and the predicted box is used to crop the search region for the next frame.
\subsection{Datasets and Settings}
\textbf{Datasets.}
\textbf{GTOT}~\cite{GTOT} is a small-scale benchmark of 50 RGB-T sequence pairs 
annotated with seven challenge attributes, widely used for rapid evaluation on 
lightweight tracking scenarios.
\textbf{RGBT210}~\cite{RGBT210} scales up to 210 sequence pairs ($\sim$210K frames) 
with 12 attribute labels, offering broader scene coverage and increased difficulty 
over GTOT.
\textbf{RGBT234}~\cite{RGBT234} extends RGBT210 with 234 sequences, including 
longer sequences of up to 8,000 frames, and shares the same 12-attribute annotation 
scheme to enable direct cross-benchmark comparison.
\textbf{LasHeR}~\cite{LasHeR} is the most comprehensive general RGB-T benchmark, 
providing 1,224 sequence pairs annotated with 19 fine-grained challenge attributes 
and covering highly diverse real-world scenarios, making it the primary testbed for 
evaluating self-supervised RGB-T trackers.
\textbf{VTUAV}~\cite{VTUAV} is a large-scale aerial benchmark consisting of 500 
UAV sequences with 1.7 million high-resolution frame pairs, targeting tracking 
under severe viewpoint variation, altitude change, and dense object distributions 
that are absent in ground-level benchmarks.
\begin{figure*}[!t]
\centering
\includegraphics[width=0.95\linewidth]{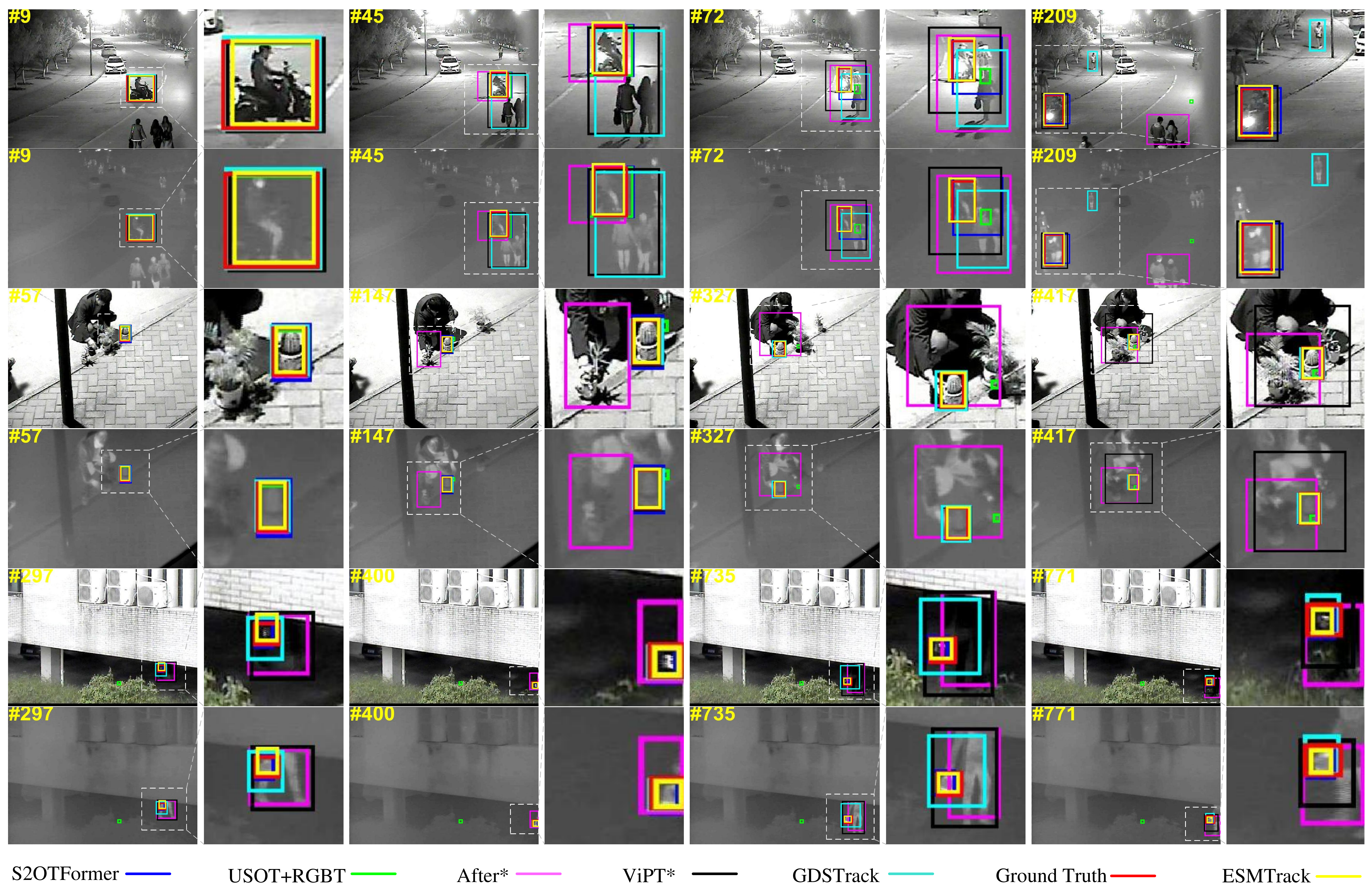}
\caption{Visualization results of ESMTrack alongside other state-of-the-art methods on the RGBT234 dataset. The top/bottom two rows depict RGB and infrared frames of \emph{elecbikechange2}, \emph{flower2} and \emph{kettle}, respectively}
\label{RGBT234-visualize-zoomin}
\end{figure*}

\textbf{Metrics.}
The performance of our tracker is evaluated using three key metrics: \textbf{Precision rate} (PR), \textbf{Normalized precision rate} (NPR), and \textbf{Success rate} (SR). PR quantifies the fraction of frames in which the Euclidean distance between the predicted object center and the ground truth center falls below a predefined threshold $\tau$. Specifically, $\tau$ is set to 5 pixels for the GTOT dataset and 20 pixels for RGBT210, RGBT234, LasHeR, and VTUAV-ST datasets. NPR is obtained by normalizing the PR.
SR measures the overlap ratio between the predicted and ground truth bounding boxes, defined as their Intersection-over-Union (IoU). It calculates the percentage of frames where the overlap exceeds a predefined threshold. For all five datasets, the final success score is computed as the area under the success rate curve (AUC).

\textbf{Implementation Details.} The proposed model is implemented using the PyTorch framework.
ViT-B256~\cite{ViT} serves as the feature encoder to facilitate interaction between the template frame and the infrared frame. The encoder is initialized with the pre-trained weights from DropMAE~\cite{DropMAE}.
The model is trained on the LasHeR dataset for 20 epochs using the AdamW optimizer with a learning rate of $2.5 \times 10^{-4}$, weight decay $10^{-4}$. The backbone learning rate is multiplied by 0.1 relative to the head. The batch size is 8. For each training sample, one template pair ($128\times128$) and three search pairs ($256\times256$) are cropped from the sequence. Two-view grounding pairs ($256\times256$) are additionally cropped from the same initial frame but with independent random jitter (center jitter factor 3, scale jitter factor 0.5), providing distinct spatial views of the annotated object for the grounding triplet loss. Triplet losses $\mathcal{L}_{tri}^{g}$ and $\mathcal{L}_{tri}^{s}$ are activated progressively: the grounding-frame triplet loss is enabled after epoch 10 and the search-frame triplet loss after epoch 15.
\subsection{Main Results}
We conducted experimental evaluations on five datasets: GTOT, RGBT210, RGBT234, LasHeR, and VTUAV-ST.
\begin{sidewaystable}
\centering
\caption{PR, NPR, and SR evaluation results compared to state-of-the-art self-supervised methods on five benchmarks. Trackers marked with \(\dagger\) denote the results obtained by directly combining RGB and infrared features using existing self-supervised RGB tracking methods. Trackers with $^\ast$ refer to the results trained with pseudo-labels using existing supervised RGB-T tracking methods. The best results are marked in \textbf{bold}, and the second-best results are \underline{underlined}}
\label{sota}
\begin{tabular}{c|c|cc|cc|ccc|cc|cc|c}
\toprule
\multirow{2}{*}{Tracker} & \multirow{2}{*}{Pub. Info.} & \multicolumn{2}{c|}{RGBT234} & \multicolumn{2}{c|}{RGBT210} & \multicolumn{3}{c|}{LasHeR} & \multicolumn{2}{c|}{VTUAV-ST} & \multicolumn{2}{c|}{GTOT} & Speed \\
 & & PR$\uparrow$ & SR$\uparrow$ & PR$\uparrow$ & SR$\uparrow$ & PR$\uparrow$ & NPR$\uparrow$ & SR$\uparrow$ & PR$\uparrow$ & SR$\uparrow$ & PR$\uparrow$ & SR$\uparrow$ & FPS$\uparrow$ \\
\midrule
USOT~\cite{USOT}$^\dagger$$^\ast$& ICCV'21  & 50.8 & 31.8 & 47.8 & 30.0 & 24.8 & 20.6 & 16.0 & 36.3 & 26.5 & 70.6 & 58.3 & 58.4 \\
TBSI~\cite{TBSI}$^\ast$        & CVPR'23  & 53.0 & 34.4 & 49.9 & 32.7 & 29.8 & 24.5 & 25.9 & 24.4 & 23.4 & 57.2 & 49.1 & 50.4 \\
ViPT~\cite{ViPT}$^\ast$        & CVPR'23  & 61.7 & 44.0 & 59.2 & 42.3 & 38.2 & 34.1 & 32.4 & 47.8 & 42.4 & 61.1 & 54.0 & $\underline{92.3}$ \\
S2OTFormer~\cite{S2OTFormer} & TOMM'24 & 68.4 & 47.7 & 65.5 & 44.5 & 39.8 & 35.4 & 29.5 & 56.7 & 44.5 & $\bm{83.1}$ & $\bm{70.2}$ & 38.2 \\
AFter~\cite{after}$^\ast$      & TIP'25   & 57.2 & 38.7 & 54.5 & 36.9 & 30.7 & 25.9 & 26.6 & 25.1 & 22.1 & 30.7 & 25.7 & 27.3 \\
GDSTrack~\cite{GDSTrack} & IJCAI'25 & $\bm{70.9}$ & $\underline{48.5}$ & $\bm{68.7}$ & $\underline{47.0}$ & $\underline{45.9}$ & $\underline{39.3}$ & $\underline{35.4}$ & $\bm{72.3}$ & $\bm{59.8}$ & 73.9 & 59.8 & 37.6 \\
\midrule
KCF~\cite{KCF}$^\dagger$   & TPAMI'15 & 46.3 & 30.5 & - & - & - & - & - & - & - & - & - & $\bm{124.1}$ \\
MEEM~\cite{MEEM}$^\dagger$  & ECCV'14  & 63.6 & 40.5 & - & - & - & - & - & - & - & 64.8 & 52.3 & 4.9 \\
UDT~\cite{UDT}$^\dagger$   & CVPR'19  & 56.8 & 42.0 & - & - & - & - & - & - & - & 73.7 & 61.4 & 74.3 \\
SSTrack~\cite{SSTrack}$^\dagger$ & AAAI'25 & 66.5 & 46.9 & 65.1 & 45.1 & 44.3 & 39.7 & 34.2 & 57.5 & 47.3 & 73.8 & 61.8 & 51.0 \\
GDSTrack (CL)~\cite{GDSTrack} & IJCAI'25 & 64.4 & 44.9 & 61.9 & 42.5 & 42.2 & 37.4 & 33.0 & 53.6 & 42.8 & 73.5 & 61.3 & 60.9\\
ESMTrack & Ours & $\underline{70.6}$ & $\bm{49.6}$ & $\underline{68.2}$ & $\bm{47.4}$ & $\bm{47.3}$ & $\bm{42.7}$ & $\bm{36.0}$ & $\underline{59.7}$ & $\underline{48.3}$ & $\underline{76.8}$ & $\underline{64.0}$ & 50.3 \\
\bottomrule
\end{tabular}%
\end{sidewaystable}

\textbf{Comparisons with state-of-the-art Trackers.}
We evaluate ESMTrack against representative trackers on five benchmarks: RGBT234, RGBT210, LasHeR, VTUAV-ST, and GTOT. Results are summarized in Table~\ref{sota}. The upper section contains pseudo-label-based methods: S2OTFormer and GDSTrack are self-supervised RGB-T trackers that generate pseudo-labels directly from multi-modal sequences; USOT$^\dagger$ extends an unsupervised RGB tracker to the RGB-T setting by incorporating the infrared modality; and TBSI$^\ast$, ViPT$^\ast$, and AFter$^\ast$ are supervised RGB-T trackers retrained using USOT-type pseudo-labels as a substitute for ground-truth annotations. The lower section contains self-supervised methods based on correlation filters or contrastive learning that do not require ground-truth annotations, including SSTrack$^\dagger$, which adapts the self-supervised RGB tracker SSTrack to the RGB-T setting by retraining with prompt-based modality fusion, and GDSTrack~(CL), which removes the temporal diffusion denoising component from GDSTrack and retrains it using the same SSTrack-type training strategy, providing a direct baseline for assessing the contribution of pseudo-label quality to GDSTrack's performance.
Among self-supervised methods in the lower section, ESMTrack consistently achieves the best performance across all five benchmarks. Compared to SSTrack$^\dagger$, ESMTrack improves PR/SR by 4.1\%/2.7\% on RGBT234, 3.1\%/2.3\% on RGBT210, 3.0\%/3.0\%/1.8\% (PR/NPR/SR) on LasHeR, 2.2\%/1.0\% on VTUAV-ST, and 3.0\%/2.2\% on GTOT, demonstrating the effectiveness of AMD and the dual triplet losses in learning discriminative RGB-T representations without additional annotations. The gains over GDSTrack~(CL) are even larger across all benchmarks, confirming that the pseudo-label supervising in GDSTrack is a critical component and that a naive contrastive learning adaptation of GDSTrack falls significantly short.

Compared to pseudo-label-based methods, ESMTrack achieves the best overall performance on \textbf{LasHeR}, surpassing GDSTrack by 1.4\%, 3.4\%, and 0.6\% in PR, NPR, and SR, respectively, and ranks first or second on all remaining benchmarks. On \textbf{RGBT234} and \textbf{RGBT210}, ESMTrack leads all methods in SR (49.6\% and 47.4\%), while attaining PR scores marginally below GDSTrack (70.6\% vs.\ 70.9\% and 68.2\% vs.\ 68.7\%). On \textbf{VTUAV-ST}, ESMTrack ranks second with PR: 59.7\% and SR: 48.3\%, outperforming all other self-supervised methods while falling behind GDSTrack, whose strong aerial performance stems from its pseudo-label-guided temporal modeling. On \textbf{GTOT}, ESMTrack achieves PR: 76.8\% and SR: 64.0\%, ranking second behind S2OTFormer, which benefits from supervised inductive biases on this small-scale benchmark.

Across all five benchmarks, ESMTrack consistently ranks among the top-2 methods and achieves the best performance on the most challenging large-scale benchmark LasHeR. Importantly, ESMTrack is trained end-to-end without offline pseudo-label generation, eliminating the additional computational overhead and error propagation inherent in two-stage training pipelines.
\begin{figure}[t]
\centering
\subfloat{\includegraphics[width=1.8in]{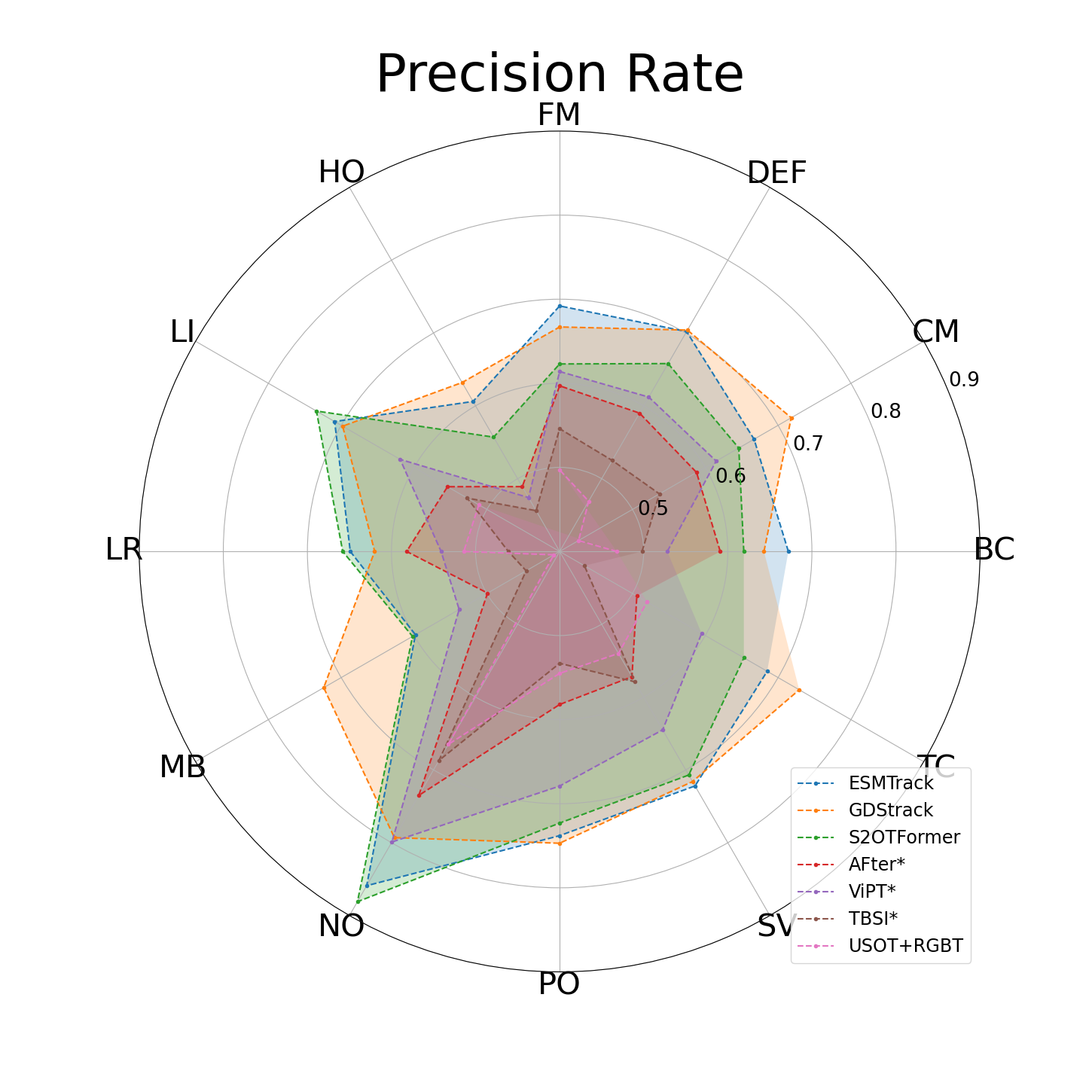}}%
\subfloat{\includegraphics[width=1.8in]{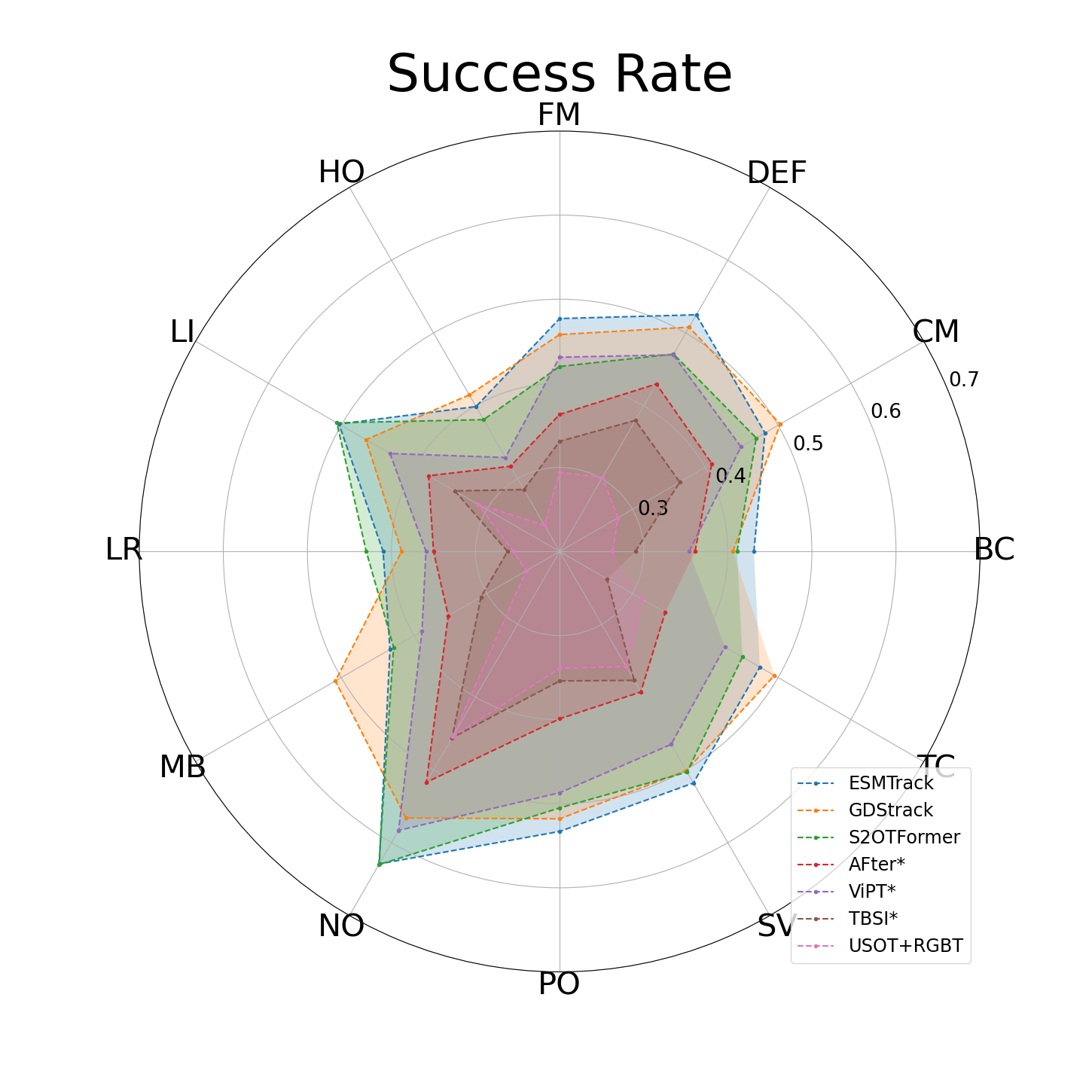}}%
\caption{Attribute-based evaluation on RGBT234 dataset. The radial axis for PR ranges from 0.4 to 0.9, while SR is scaled from 0.2 to 0.7}
\label{RGBT234-attr}
\end{figure}
\begin{table*}[t]
\caption{PR/SR evaluation results on Attribute Challenges comparing to state-of-the-art methods on the LasHeR dataset}
\centering
\resizebox{\textwidth}{!}{
\begin{tabular}{c c c c c c c c c} 
 \toprule
    Tracker & USOT+RGBT & AFter$^\ast$ & ViPT$^\ast$ & TBSI$^\ast$& S2OTFormer & GDSTrack &  ESMTrack\\
   \midrule
   NO & 52.7/38.2 & 56.2/45.0 & \underline{72.4}/\underline{56.5} & 60.3/46.6 &67.0/48.8 & 67.8/52.1 & $\bm{78.0}/\bm{57.3}$\\
   PO & 20.4/12.6& 27.6/24.4 & 33.5/29.4 & 25.6/23.2 & 35.3/26.4 & \underline{42.6}/$\bm{33.0}$ &$\bm{42.8}$/\underline{32.8} \\
   TO & 15.4/9.6 & 30.2/25.5 & 29.4/26.3 & 23.6/21.2 & 35.0/25.4 & $\bm{42.2}/\bm{32.1}$ & \underline{41.2}/\underline{31.0}\\
   HO  & 6.2/4.8 & 18.9/25.7 & 15.4/21.2 & 16.4/24.9 & 20.3/19.1& $\bm{34.6}/\bm{36.1}$  & \underline{31.6}/\underline{31.7}\\
   OV & 20.5/16.9 & 16.9/22.4 & $\bm{55.0}$/$\bm{51.4}$ & 25.2/28.3 & 30.6/28.7 & 50.1/41.8 & \underline{52.4}/\underline{46.7}\\
   LI & 15.3/10.2 & 30.8/26.7 & 30.7/26.9 & 25.7/23.5 & 33.3/25.1 & $\bm{36.7}/\bm{29.2}$ &\underline{36.2}/\underline{28.0} \\
   HI & 35.8/24.2 & 37.4/29.7 & 46.7/36.2 & 37.9/30.4 & 46.1/34.3& \underline{53.0}/\underline{39.4}  & $\bm{54.7}/\bm{40.6}$\\
   AIV  & 4.0/3.8 & $\bm{29.4}$/$\bm{30.7}$ & \underline{27.2}/\underline{27.9} & 23.1/25.4 & 19.5/18.0& 21.9/21.2  & 22.5/21.0\\
   LR & 23.5/13.2 & 25.8/17.7 & 29.3/21.8 & 23.9/15.8 & 36.3/24.4 & \underline{40.6}/$\bm{27.1}$ & $\bm{40.8}$/\underline{26.8}\\
   DEF & 16.8/12.2 & 39.7/35.7 & 43.0/$\bm{38.7}$ & 34.6/31.8 & \underline{44.8}/34.6 & $\bm{47.4}$/\underline{37.9} & $\bm{47.4}$/37.5\\
   BC  & 17.9/12.3 & 30.6/26.7 & 36.7/31.7 & 27.9/25.3 & 36.1/27.7 & \underline{41.9}/\underline{33.3} & $\bm{44.9}/\bm{34.1}$\\
   SA & 18.6/11.2 & 21.0/21.5 & 29.7/27.3 & 20.0/21.0 & 31.3/23.5 & $\bm{39.5}/\bm{30.8}$ & \underline{38.2}/\underline{30.0}\\
   TC & 20.0/12.1 & 22.9/20.9 & 29.3/25.5 & 23.0/20.9 & 32.4/24.0 & $\bm{40.4}/\bm{30.7}$ & \underline{37.6}/\underline{28.4}\\
   MB & 19.5/11.8 & 23.4/21.5 & 31.4/27.2 & 20.7/19.8 & 32.2/23.6 & $\bm{45.1}/\bm{34.3}$ & \underline{38.8}/\underline{29.2}\\
   CM & 21.5/13.7 & 27.6/24.2 & 36.5/30.7 & 24.1/22.3 & 36.9/27.5 & $\bm{47.9}/\bm{36.5}$ & \underline{43.4}/\underline{32.7}\\
   FL  & 15.0/5.9 & 22.4/20.0 & 28.2/25.1 & 21.8/19.0 & 25.8/18.4 & $\bm{48.5}/\bm{34.9}$ & \underline{34.0}/\underline{26.5}\\
   FM & 21.0/14.2 & 29.1/26.4 & 37.2/32.4 & 27.8/25.7  & 35.7/27.5 & \underline{44.2}/\underline{35.0} & $\bm{45.8}/\bm{35.6}$\\
   SV & 24.6/15.9 & 31.3/27.1 & 38.6/32.9 & 30.6/26.6 & 38.9/29.0 & \underline{46.1}/\underline{35.6} & $\bm{47.1}/\bm{36.1}$\\
   ARC  & 16.2/10.8 & 27.7/26.1 & 34.0/30.8 & 26.6/25.1 & 32.7/25.8 & \underline{41.6}/\underline{33.7} & $\bm{43.8}$/$\bm{34.9}$\\
 \bottomrule
\end{tabular}
}
\label{attributes-LasHeR}
\end{table*}

\textbf{Attribute Analysis.}
Figure~\ref{RGBT234-attr} presents the per-attribute performance of ESMTrack and competing methods on RGBT234 across twelve challenge factors. ESMTrack achieves the top ranking in BC, FM, SV, PO, and DEF under both PR and SR metrics, demonstrating strong generalization across diverse challenge conditions. In background clutter (BC), the adaptive modality 
decoupling mechanism suppresses interference from the dominant modality when it is corrupted by distractors, preventing unreliable activations from polluting the fused representation. Under fast motion (FM) and scale variation (SV), the cross-frame temporal triplet loss provides consistent supervision across frames with large displacement or appearance variation, contributing to robust localization without bounding box annotations. In 
deformation (DEF) and partial occlusion (PO), the grounding triplet loss anchors training on initial templates and enforces spatial discriminability against shifted negatives, enabling the model to maintain object identity under appearance changes.
In low illumination (LI) and no occlusion (NO), ESMTrack ranks second, marginally behind S2OTFormer, while remaining competitive with other self-supervised counterparts. The primary performance gap appears in motion blur (MB) and thermal crossover (TC), where GDSTrack leads by a notable margin. Motion blur degrades response map quality, which reduces the reliability of APCE-based modality importance estimation and weakens the gating signal. 
Thermal crossover introduces ambiguity between IR and RGB cues, posing a fundamental challenge for modality-aware fusion under self-supervised training. These attributes suggest directions for future improvement, such as more robust response quality estimation under degraded observations.

\begin{figure*}[t]
\centering
\includegraphics[width=\linewidth]{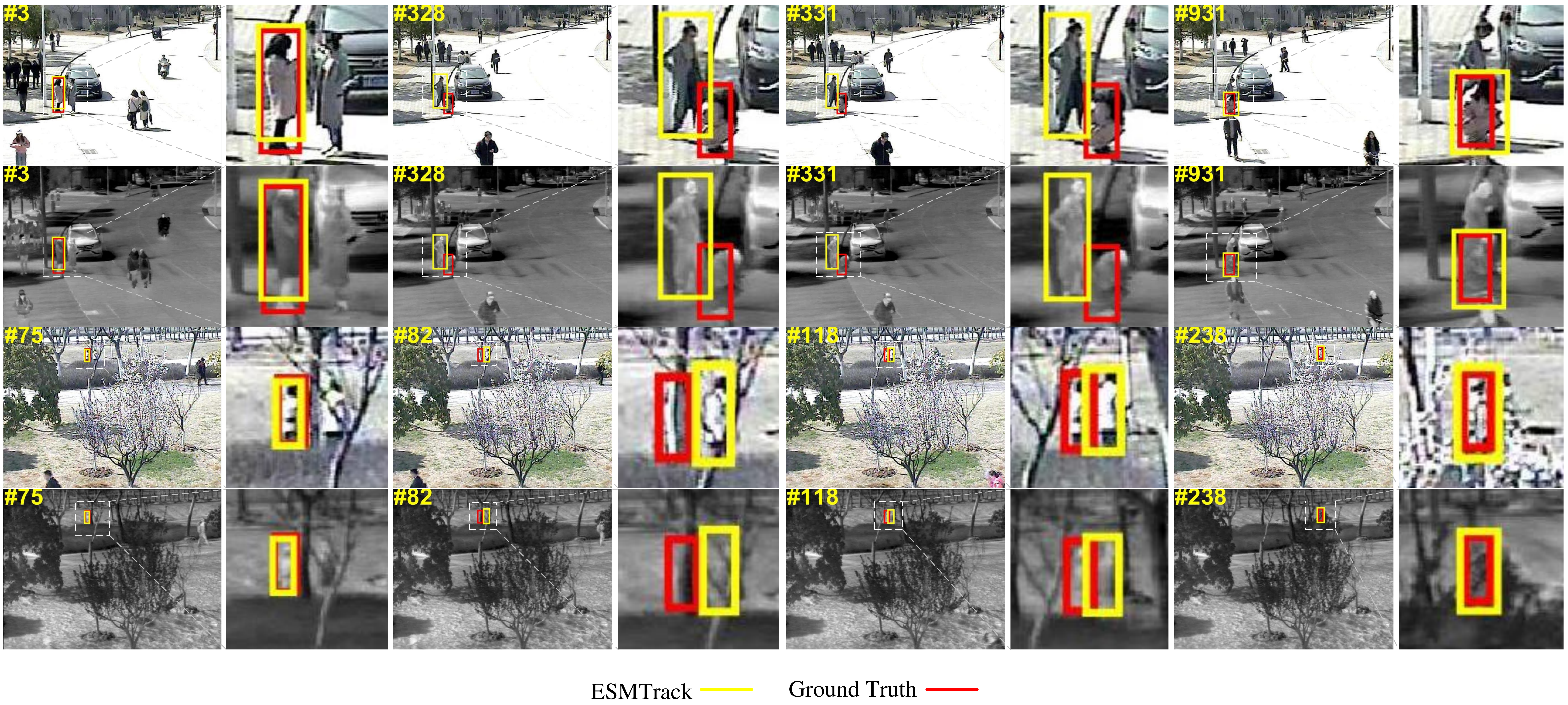}
\caption{Failure cases on sequence \emph{crouch} and \emph{diamond} in RGBT234 dataset. The first row shows RGB images, and the second row shows the corresponding thermal images}
\label{FailureCases-roomin}
\end{figure*}

Table~\ref{attributes-LasHeR} reports per-attribute PR/SR performance on LasHeR. ESMTrack achieves the best results on nine attributes and ranks second on most of the remaining ones, demonstrating broad robustness across diverse challenge conditions. ESMTrack leads on motion-related attributes including fast motion (FM: 45.8\%/35.6\%), scale variation (SV: 47.1\%/36.1\%), and aspect ratio change (ARC: 43.8\%/34.9\%), benefiting from the cross-frame temporal triplet loss, which enforces consistent target representations across frames with large displacement and appearance variation. Under stable illumination conditions, the APCE-based modality gating correctly identifies the more reliable modality, yielding the best performance on no occlusion (NO: 78.0\%/57.3\%) and high illumination (HI: 54.7\%/40.6\%). The AMD module's ability to suppress background interference further contributes to the best results on background clutter (BC: 44.9\%/34.1\%). GDSTrack outperforms ESMTrack on heavy occlusion (HO), total occlusion (TO), motion blur (MB), camera motion (CM), and fast-and-long-term (FL) attributes by notable margins, particularly on FL (34.0\%/26.5\% vs. 48.5\%/34.9\%). These gaps are consistent with the failure cases discussed in Figure~\ref{FailureCases-roomin}: without an explicit re-detection mechanism, accumulated drift under prolonged occlusion or long-term fast motion cannot be corrected. Motion blur degrades response map quality and weakens the APCE estimation, reducing the reliability of the modality gating signal. Under abrupt illumination variation (AIV), ESMTrack ranks third behind AFter$^\ast$ (29.4\%/30.7\%) and ViPT$^\ast$ (27.2\%/27.9\%), as sudden illumination changes violate the smoothness assumption implicit in APCE-based importance estimation.

\textbf{Speed analysis.}
As shown in Table~\ref{sota}, ESMTrack achieves 50.3 fps, surpassing the pseudo-label-based methods S2OTFormer (38.2 fps) and GDSTrack (37.6 fps) by 31.7\% and 33.8\%, respectively. Beyond the speed advantage, our method eliminates the two-stage training pipeline required by pseudo-label approaches, where offline label generation introduces additional computational overhead and potential error propagation from imperfect pseudo-labels. ESMTrack is trained end-to-end directly on raw RGB-T data, making it both more efficient at inference and simpler to train.
\begin{figure}[h]
\centering
\subfloat{\includegraphics[width=1.7in]{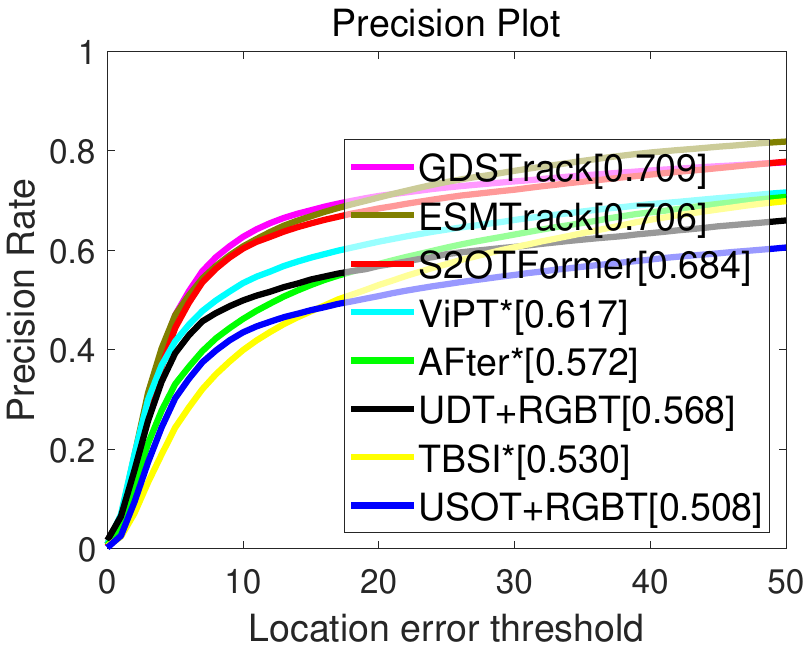}}%
\subfloat{\includegraphics[width=1.7in]{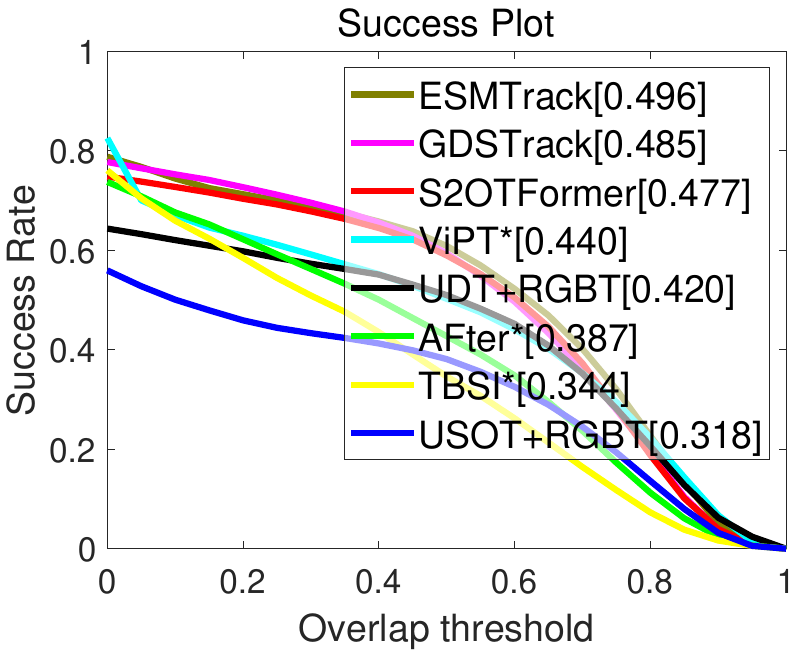}}%
\caption{Precision Rate and Success Rate on RGBT234 dataset compared against other RGB-T trackers}
\label{RGBT234-PR-SR}
\end{figure}

\begin{figure}[h]
\centering
\subfloat{\includegraphics[width=1.7in]{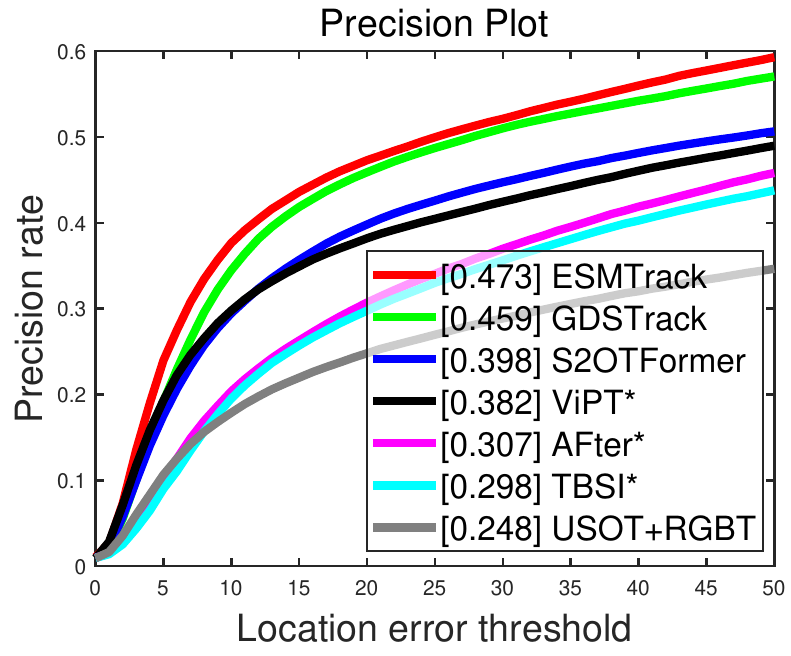}}%
\subfloat{\includegraphics[width=1.7in]{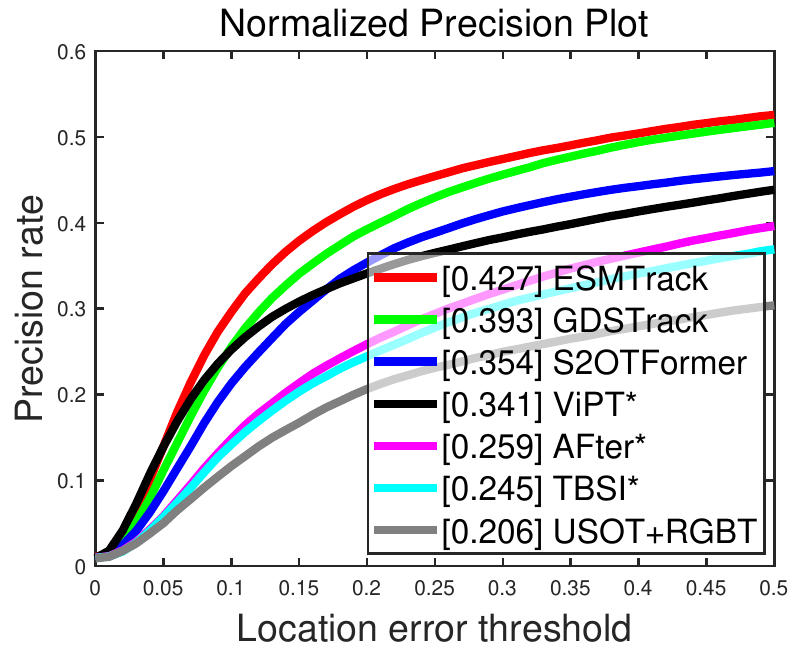}}%
\subfloat{\includegraphics[width=1.7in]{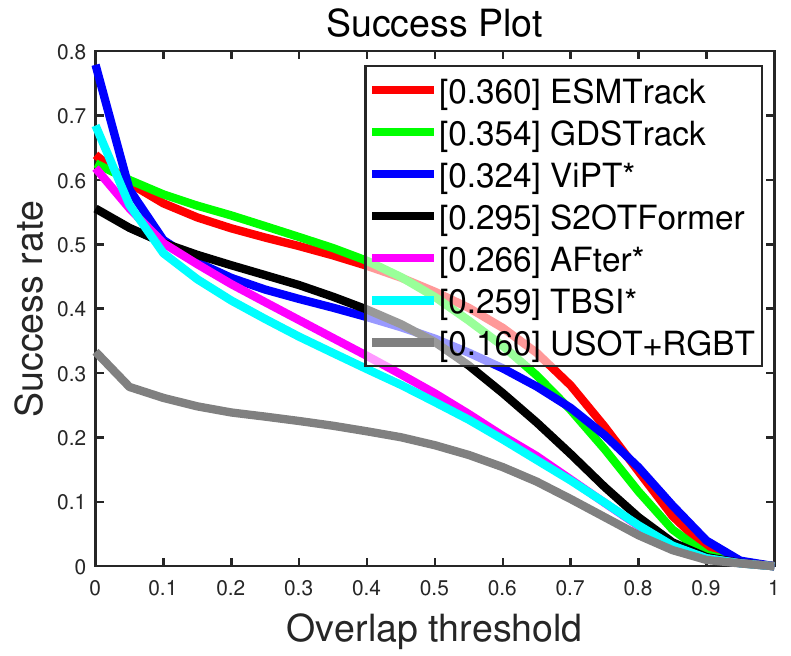}}%
\caption{Precision Rate, Normalized Precision Rate and Success Rate on LasHeR dataset compared against other RGB-T trackers}
\label{LasHeR-PR-SR}
\end{figure}
\textbf{Analysis of $PR$, $NPR$, and $SR$ plots.}
We compare our model with state-of-the-art RGB-T trackers on the RGBT234 dataset. As shown in Figure~\ref{RGBT234-PR-SR}, our model achieves the second-best precision rate (PR) and the best success rate (SR) among all compared methods.

We compare our model with state-of-the-art RGB-T trackers on the LasHeR dataset. As shown in Figure~\ref{LasHeR-PR-SR}, our model achieves the best performance in both precision rate (PR), Normalized precision rate (NPR) and success rate (SR) plots.

\textbf{Tracking Visualization Analysis.}
Figure~\ref{RGBT234-visualize-zoomin} presents qualitative comparisons on three challenging sequences from RGBT234: \emph{elecbikechange2}, \emph{flower2}, and \emph{kettle}, shown in paired RGB and infrared frames.
In the \emph{elecbikechange2} sequence, the object undergoes substantial appearance variation as the electric bike changes its configuration under varying illumination conditions. From frame \emph{\#9} to \emph{\#209}, most competing methods progressively accumulate drift, with several trackers 
shifting to nearby vehicles sharing similar thermal signatures in the infrared modality. 
In the \emph{flower2} sequence, dense background clutter from foliage creates ambiguous texture and thermal responses that challenge all methods. As the object moves among the cluttered background, competing trackers exhibit increasing localization errors from frame \emph{\#147} onward, with most methods producing significantly misaligned boxes by frame \emph{\#417}. 
In the \emph{kettle} sequence, the object is small and undergoes scale variation alongside partial occlusion. By frame \emph{\#735}, most competing methods have drifted entirely off the object, with some producing severely displaced predictions.

\begin{figure}[t]
    \centering
    \includegraphics[width=\linewidth]{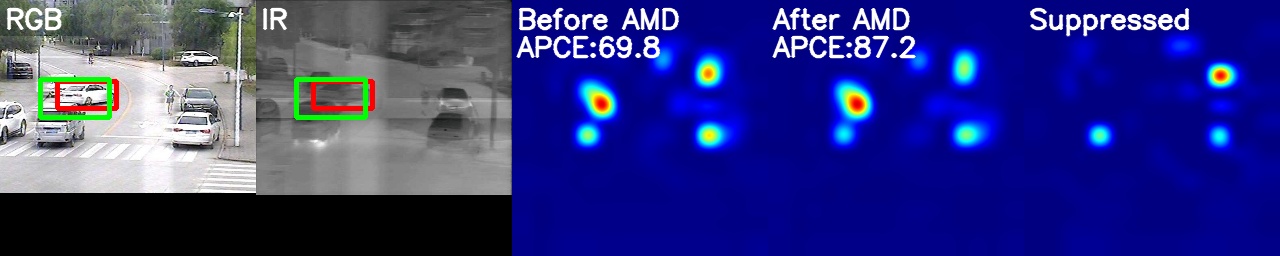}
    \vspace{0.1mm}
    \includegraphics[width=\linewidth]{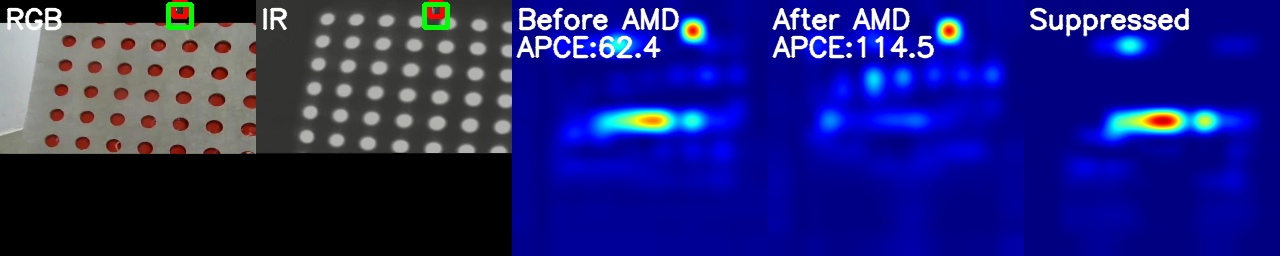}
    \vspace{0.1mm}
    \includegraphics[width=\linewidth]{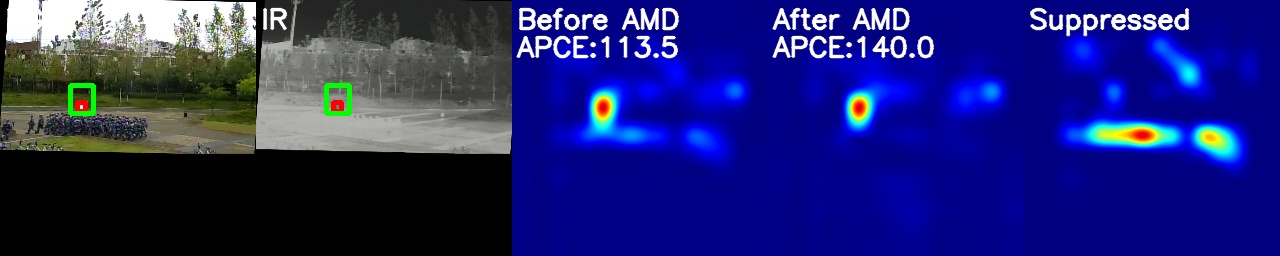}
    \vspace{0.1mm}
    \includegraphics[width=\linewidth]{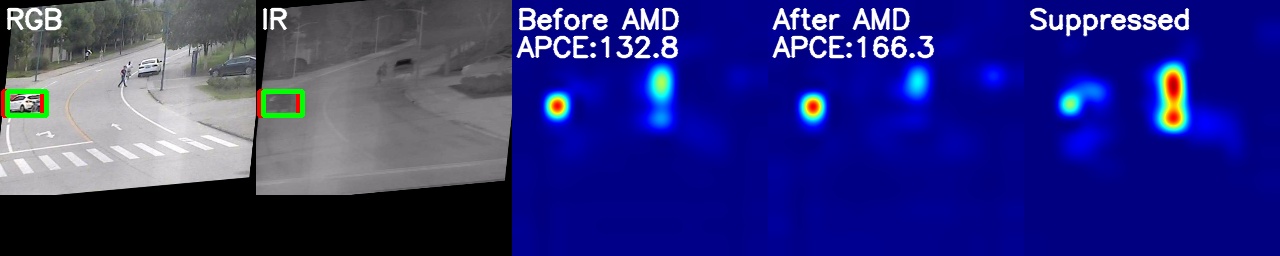}
    \vspace{0.1mm}
    \includegraphics[width=\linewidth]{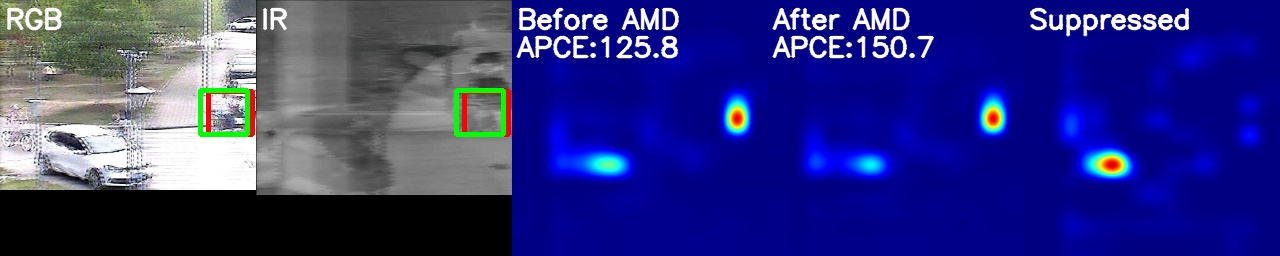}
    \caption{Visualization of the Adaptive Modality Decoupling (AMD) module on five representative sequences from the LasHeR dataset.}
    \label{AMD-response}
\end{figure}

\textbf{Visualization Analysis of AMD Module.}
Figure~\ref{AMD-response} presents five representative tracking scenarios from the LasHeR dataset, each visualizing the RGB and IR search frames alongside three response maps: the fused branch response before AMD suppression \textbf{(Before AMD)}, after AMD suppression \textbf{(After AMD)}, and the suppressed content \textbf{(Suppressed)}. As shown, the APCE metric consistently increases after AMD across all five scenarios, demonstrating that AMD reliably produces sharper and more discriminative target responses. The Suppressed panels further reveal that AMD predominantly removes diffuse background activations and distractor responses rather than target-relevant information, confirming that the module effectively identifies and decouples redundant inter-modal features from the fused representation. These visualizations validate that AMD promotes complementary feature learning between modalities, leading to more localized and robust target responses across diverse and challenging tracking conditions.
\subsection{Ablation Study}
We conducted ablation studies on GTOT, RGBT234, RGBT210, and LasHeR datasets to validate the effectiveness of modules we proposed.

\textbf{Component Analysis.}
Table~\ref{Ab-pipeline} reports the contribution of each proposed component through stepwise ablation across four benchmarks. Adding the AMD module yields clear gains on GTOT (PR: +4.5\%, SR: +4.1\%) and RGBT234 (PR: +4.1\%, SR: +3.5\%). This substantial gain confirms that task-driven modality importance estimation is critical in the absence of ground-truth supervision: by suppressing the residual contribution of the dominant modality, \textbf{AMD} reduces shortcut learning and improves robustness to modality degradation. The moderate gains on RGBT210 (PR: +1.8\%, SR: +1.7\%) and LasHeR (PR: +3.7\%, NPR: +3.4\%, SR: +3.7\%) further demonstrate its consistent effectiveness across scenes of varying complexity.
Incorporating the grounding triplet loss~\textbf{(Ground. Tri.)} on top of AMD brings further improvements across all benchmarks, with the largest gains on RGBT210 (PR: +2.0\%, SR: +0.8\%) and LasHeR (PR: +2.2\%, NPR: +2.6\%, SR: +2.1\%), where scene diversity is high. The grounding triplet loss anchors the representation on annotated initial frames and enforces spatial discriminability between the object and shifted negatives, providing explicit localization supervision that complements the AMD gating signal.
Adding the temporal triplet loss~\textbf{(Temporal Tri.)} for search frames further improves performance on GTOT (PR: +1.3\%, SR: +1.6\%) and RGBT234 (PR: +0.4\%, SR: +1.1\%), demonstrating that forward-backward consistency filtering provides reliable temporal supervision for unlabeled search frames. The marginal trade-offs observed on RGBT210 (PR: $-$0.9\%) and LasHeR (SR: $-$0.9\%) suggest that hard negative mining from batch samples occasionally introduces conflicting gradients on sequences with high inter-instance similarity, which we identify as a direction for future refinement.

\begin{table}[t]
\caption{PR, NPR, and SR of the model with different module}
\centering
\begin{tabular}{c c c c c c c} 
 \toprule
 AMD&Ground. Tri.&Temporal Tri. & GTOT & RGBT234 & RGBT210 & LasHeR\\
   \midrule
 && & 72.1/59.3 & 64.1/43.9 &63.6/43.4 & 40.4/36.1/30.2\\
 \checkmark&&  & 76.6/63.4 & 68.2/47.4&65.4/45.1& 44.1/39.5/33.9\\ %
 \checkmark&\checkmark&& 78.0/63.5 & 68.7/47.4 &$\bm{67.4}$/45.9& 46.3/42.1/$\bm{36.0}$\\
 \checkmark&\checkmark&\checkmark& $\bm{79.3}/\bm{65.1}$ & $\bm{69.1}/\bm{48.5}$ &66.5/$\bm{46.3}$&$\bm{47.0}/\bm{42.5}$/35.1\\
 \bottomrule
\end{tabular}
\label{Ab-pipeline}
\end{table}
\begin{table}[t]
\caption{Ablation on trainable parameter configurations. ``$\dagger$'' denotes the final adopted setting}
\centering
\begin{tabular}{ c c c c c} 
 \toprule
 Trainable Parameters & GTOT & RGBT234 & RGBT210 & LasHeR\\
 \midrule
 Task-specific modules only & $\bm{79.3}/\bm{65.1}$ & 69.1/48.5 & 66.5/46.3 & 47.0/42.5/35.1\\
 + LayerNorm \& CLS token$^\dagger$ & 76.8/64.0 & $\bm{70.6}/\bm{49.6}$ & $\bm{68.2}/\bm{47.4}$ & $\bm{47.3}/\bm{42.7}$/$\bm{36.0}$ \\
 \bottomrule
\end{tabular}
\label{tab:learn-para}
\end{table}
\begin{table}[t]
\caption{PR, NPR, and SR of the model with different modality suppression methods}
\centering
\begin{tabular}{ c c c c c} 
 \toprule
 Modality Suppression & GTOT & RGBT234 & RGBT210 & LasHeR\\
   \midrule
 w/o Suppression & 72.2/60.0 & 62.3/43.7 & 60.8/41.9 & 42.0/38.6/32.4 \\
 Random Suppression & 75.7/63.3 & 63.8/44.9 & 61.7/42.6 & 44.0/39.9/33.1 \\
 AMD & $\bm{79.3}/\bm{65.1}$ & $\bm{69.1}/\bm{48.5}$ & $\bm{66.5}/\bm{46.3}$ & $\bm{47.0}/\bm{42.5}/\bm{35.1}$\\
 \bottomrule
\end{tabular}
\label{AMD-ab}
\end{table}
\begin{table}[t]
\caption{Ablation on response-map reliability metrics for the AMD module}
\centering
\begin{tabular}{ c c c c c} 
 \toprule
 Reliability Metric & GTOT & RGBT234 & RGBT210 & LasHeR\\
 \midrule
 Negative Entropy & $\bm{77.5}/\bm{64.1}$ & 64.8/45.4 & 63.9/44.6 & 44.5/40.1/34.1 \\
 Max Response & 68.3/56.9 & 67.3/46.7 & 65.4/45.0 & 44.6/40.1/34.9\\
 APCE& 76.8/64.0 & $\bm{70.6}/\bm{49.6}$ & $\bm{68.2}/\bm{47.4}$ & $\bm{47.3}/\bm{42.7}$/$\bm{36.0}$ \\
 \bottomrule
\end{tabular}
\label{reliability}
\end{table}
\begin{table}[t]
\caption{PR, NPR, and SR of the model with different negative sampling strategies for triplet loss}
\centering
\begin{tabular}{c c c c c c} 
 \toprule
 Grounding Tri. & Temporal Tri. & GTOT & RGBT234 & RGBT210 & LasHeR\\
  \midrule
  Cross Sequence & Shifted Box & 75.2/62.3 & 63.2/44.1 & 62.0/42.8 & 42.8/37.9/32.8\\
  Cross Sequence & Cross Sequence& $\bm{80.3}/\bm{66.2}$ & 67.5/47.0 & 65.3/44.8 & 44.6/39.8/33.6 \\
  \midrule
  Shifted Box & Shifted Box & 74.5/61.4 & 62.5/43.6 & 58.1/40.3 & 43.6/39.2/32.6\\
  Shifted Box & Cross Sequence& 79.3/65.1 & $\bm{69.1}/\bm{48.5}$ & $\bm{66.5}/\bm{46.3}$ & $\bm{47.0}/\bm{42.5}/\bm{35.1}$ \\
  \bottomrule
\end{tabular}
\label{triplet-loss-neg}
\end{table}
\begin{table}[t]
\caption{PR, NPR, and SR of the model with different pseudo-label filtering strategy}
\centering
\begin{tabular}{ c c c c c} 
 \toprule
 Temporal Triplet Loss & GTOT & RGBT234 & RGBT210 & LasHeR\\
   \midrule
 w/o Filter& 74.0/61.2 & 65.2/46.0 & 64.4/44.8 & 44.7/40.5/33.8\\
 Forward APCE Filter & 75.8/62.6 & 67.8/47.5 & 66.3/45.8 & $\bm{47.2}/\bm{42.6}/\bm{35.7}$ \\
 Backward IOU Filter& $\bm{79.3}/\bm{65.1}$ & $\bm{69.1}/\bm{48.5}$ &$\bm{66.5}/\bm{46.3}$& 47.0/42.5/35.1\\
 \bottomrule
\end{tabular}
\label{search-filter}
\end{table}
\begin{table}[t]
\caption{PR, NPR, and SR of the model with different $\mathcal{L}_{rgb\text{-}tir}$ settings}
\centering
\begin{tabular}{c c c c c c} 
 \toprule
 Cross Modal.&Cross View& GTOT & RGBT234 & RGBT210 & LasHeR\\
   \midrule
 && 74.4/61.8 & 61.3/42.8 & 60.3/41.6 & 43.0/38.9/32.4\\
 \checkmark&& 71.2/59.7 & 63.9/44.6 &63.2/43.4 &43.2/39.1/32.5 \\
 \checkmark&\checkmark& $\bm{79.3}/\bm{65.1}$ & $\bm{69.1}/\bm{48.5}$ &$\bm{66.5}/\bm{46.3}$&$\bm{47.0}/\bm{42.5}/\bm{35.1}$\\
 \bottomrule
\end{tabular}
\label{Ab-rgb-ir-loss}
\end{table}

\textbf{Effect of Trainable Parameter Configuration.}
Table~\ref{tab:learn-para} compares two trainable parameter configurations while keeping the shared backbone layers frozen. The first strategy trains only task-specific modules, and serves as the basis for all other ablation experiments. The second strategy additionally unfreezes the backbone 
normalization layers and the CLS token.
Training task-specific modules only yields higher scores on GTOT, where the limited dataset scale makes full adaptation of normalization statistics unnecessary. When normalization layers and the CLS token are additionally trained, performance improves consistently across RGBT234, RGBT210, and LasHeR. This suggests that the distribution of RGB-T features differs non-trivially from the single-modality pretraining distribution, and allowing the backbone to recalibrate its feature normalization statistics 
facilitates better adaptation to the dual-modality input. The CLS token, serving as a global context summary, further benefits from being task-specifically tuned when the input modality composition changes. We therefore adopt the second configuration as the final model.

\textbf{The ablation of criterion for modality suppression in AMD.}
As shown in Table~\ref{AMD-ab}, we conducted ablation studies to evaluate the effect of modality decoupling. As a baseline, we first consider the setting without any suppression~\textbf{(w/o Suppression)}, where fused features are fed directly into the tracking head; results are reported in the first row. We then compare two dropout strategies: Random Suppression~\textbf{(Random Suppression)}, which removes one modality from the fusion branch at random, and our proposed Adaptive Modality Decoupling strategy~\textbf{(AMD)}, which subtracts a weighted modality from the fusion branch based on estimated modality importance (third row).
Random suppression consistently outperforms the no-dropout baseline, demonstrating that modality redundancy persists after fusion in self-supervised RGB-T tracking and confirming the general effectiveness of dropout-based regularization. Building on this, AMD further improves upon Random Drop by explicitly evaluating modality importance, enabling more targeted suppression of redundant information.
Quantitatively, compared with Random suppression, AMD achieves gains of 3.6\% in PR and 1.8\% in SR on GTOT, 5.3\% in PR and 3.6\% in SR on RGBT234, and 4.8\% in PR and 3.7\% in SR on RGBT210. On the more challenging LasHeR dataset, AMD improves PR, NPR, and SR by 3.0\%, 2.6\%, and 2.0\%, respectively.

\textbf{Effect of Reliability Metric in AMD.}
We compare three response-map metrics for modality reliability estimation in AMD: APCE, Max Response, and Negative Entropy. As shown in Table~\ref{reliability}, \textbf{APCE} achieves the best results on RGBT234, RGBT210, and LasHeR, while negative entropy performs comparably on the simpler GTOT benchmark. \textbf{Max Response} degrades substantially on GTOT due to response saturation: when both modalities yield high-confidence responses, the inter-modality delta collapses to near zero, reducing AMD to uniform fusion. \textbf{Negative Entropy} fails on complex benchmarks because multi-distractor scenes can produce multi-peaked response maps with lower entropy than single-peaked ones, yielding an inverted reliability signal. In contrast, APCE measures peak-to-noise ratio rather than absolute peak height or distributional shape, providing a scale-invariant and clutter-aware reliability estimate that generalizes across benchmark difficulty levels. We therefore adopt APCE as the reliability metric in all experiments.

\textbf{Effect of Negative Sampling Strategies for Triplet Loss.}
We investigate the impact of different negative sampling strategies for grounding and temporal triplet losses across four benchmarks. Results are presented in Table~\ref{triplet-loss-neg}. Specifically, \textbf{Cross Sequence} negatives are sampled from a different video sequence, providing semantically distinct distractors; \textbf{Shifted Box} negatives are spatial crops from the same frame with the target region displaced: for the grounding triplet, the crop is centered on a box shifted from the ground-truth annotation of the initial frame; for the temporal triplet, the crop is centered on a box shifted from the forward-tracking pseudo-label of the search frame. This provides hard in-sequence distractors that share scene context with the anchor.

Cross Sequence search negatives consistently outperform Shifted search negatives.
Across all grounding strategies and all four benchmarks, replacing Shifted search negatives with Cross Sequence negatives yields consistent improvements. When paired with cross-sequence grounding, replacing shifted search negatives with cross-sequence negatives improves performance on GTOT from (PR: 75.2\%, SR: 62.3\%) to (PR: 80.3\%, SR: 66.2\%) and on RGBT234 from (PR: 63.2\%, SR: 44.1\%) to (PR: 67.5\%, SR: 47.0\%). A similar trend holds when paired with Shifted grounding. This consistent gap reveals a fundamental limitation of applying spatial-shifted negatives to search frames: unlike grounding frames whose anchors are derived from ground-truth annotations, search frame anchors rely on predicted bounding boxes. Spatial discrimination applied to such imprecise anchors introduces noisy gradients that destabilize learning, regardless of the grounding configuration.
\begin{table}
\centering
\caption{PR, NPR, and SR on the backward IoU threshold for temporal triplet loss sample filtering}
\begin{tabular}{c c c c c c}
 \toprule
  Backward IOU & 0.1 & 0.2 & 0.3 & 0.4\\
  \midrule
  GTOT  & 77.6/63.2 &73.6/59.6& $\bm{79.3}/\bm{65.1}$ & 77.2/62.9\\
  RGBT234 & 66.5/46.5 &64.5/45.3& $\bm{69.1}/\bm{48.5}$ &67.9/47.4\\
  LasHeR & 44.4/39.3/33.4 &42.8/38.3/32.2& $\bm{47.0}/\bm{42.5}/\bm{35.1}$ & 44.6/40.0/33.7\\
 \bottomrule
\end{tabular}
\label{backward_iou}
\end{table}

The choice of grounding negative strategy is dataset-dependent.
Among the two cross-sequence search configurations, cross-sequence grounding achieves the highest performance on GTOT (PR: 80.3\%, SR: 66.2\%), while shifted grounding consistently yields better results on RGBT234 (PR: 69.1\%, SR: 48.5\% vs. PR: 67.5\%, SR: 47.0\%), RGBT210 (PR: 66.5\%, SR: 46.3\% vs. PR: 65.3\%, SR: 44.8\%), and LasHeR (PR: 47.0\%, NPR: 42.5\%, SR: 35.1\% vs. PR: 44.6\%, NPR: 39.8\%, SR: 33.6\%). GTOT is a relatively small and simple benchmark where sequences are well-separated; cross-sequence identity discrimination alone provides sufficient discriminative pressure. In contrast, RGBT234, RGBT210, and LasHeR contain more diverse sequences with visually similar objects, where spatial-shifted grounding negatives provide an additional fine-grained localization signal that generalizes better.

Complementary strategies yield the best overall performance.
The combination of Shifted grounding and Cross Sequence search negatives achieves the best aggregate performance across benchmarks. These two objectives are orthogonal: the grounding triplet loss, with reliable GT-based anchors, enforces spatial precision by penalizing nearby incorrect locations; the temporal triplet loss enforces cross-instance identity discrimination without requiring precise spatial anchors. This complementarity avoids gradient redundancy and allows the model to jointly optimize for localization accuracy and cross-sequence discriminability. Both homogeneous configurations where grounding and search frames share the same negative sampling strategy are outperformed by the proposed asymmetric design. Notably, Shifted+Shifted achieves the lowest overall performance, further validating the necessity of aligning the negative construction strategy with the reliability of the anchor source.

\textbf{The ablation of Pseudo-Label filtering strategy.}
During the computation of the temporal triplet loss, the anchor is obtained by pooling pseudo-labels generated via forward tracking. Unlike prior pseudo-label-based methods that directly supervise regression with generated boxes, these pseudo-labels serve solely as anchor features for contrastive learning, so label noise affects representation quality rather than causing direct regression errors. However, due to the lack of ground-truth supervision in forward tracking, these pseudo-labels inevitably contain noise. To address this issue, we leverage the IoU between backward tracking results and ground truth to filter pseudo-labels, as backward predictions can intuitively reflect the quality of tracking in the search frame \textbf{(Backward IOU Filter)}. We also conduct comparative experiments without pseudo-label filtering \textbf{(w/o Filter)} and with filtering based on APCE of the forward score map \textbf{(Forward APCE Filter)}, as reported in Table~\ref{search-filter}. The results confirm that enforcing temporal consistency via backward tracking provides a more reliable criterion for pseudo-label selection than confidence-based heuristics. Taking RGBT234 as an example, using backward IoU filtering improves PR and SR by 3.9\% and 2.5\%, respectively, compared to the setting without pseudo-label filtering. Compared with forward score map-based filtering, it further improves PR and SR by 1.3\% and 1.0\%, respectively.

\textbf{The ablation of $\mathcal{L}_{rgb\text{-}tir}$.}
To validate the effectiveness of the proposed RGB-TIR cross-view contrastive loss $\mathcal{L}_{rgb\text{-}tir}$, we conduct an ablation in which this term is removed from the full model; results are reported in the first row of Table~\ref{Ab-rgb-ir-loss}. We further observe that jointly applying both the cross-modal and cross-view losses yields consistently stronger performance than relying on the cross-modal loss alone.
Specifically, the combined setting improves PR and SR by 8.1\% and 5.4\% on GTOT, 5.2\% and 3.9\% on RGBT234, and 3.3\% and 2.9\% on RGBT210. On the more challenging LasHeR dataset, PR, NPR, and SR are improved by 3.8\%, 3.4\%, and 2.6\%, respectively.
\subsection{Failure Cases}
We identify two distinct failure modes from our analysis of \emph{crouch} and \emph{diamond} on RGBT234, as shown in Figure~\ref{FailureCases-roomin}.
In \emph{crouch}, the tracker maintains accurate localization in frame \emph{\#3}, but progressive scale variation gradually degrades tracking quality. As the object undergoes significant changes in scale, the APCE-based modality importance estimation becomes less reliable, since both RGB and IR response maps deteriorate simultaneously under extreme scale compression. By frame \emph{\#328}, a noticeable offset between the predicted box and ground truth emerges, which further accumulates into severe drift at frame \emph{\#331}. This case reveals a limitation of our response-quality-based fusion gating: when object scale changes are drastic and sustained, the gating signal loses its discriminative power across both modalities, leaving the tracker without a reliable cue for self-correction.
In \emph{diamond}, the tracker handles well in frame \emph{\#75}, but occlusion introduced by foreground structures causes abrupt appearance disruption. At frame \emph{\#82}, partial occlusion degrades both the RGB texture and the infrared thermal signature of the object, weakening the cross-modal fusion signal. As occlusion deepens at frame \emph{\#118}, the tracker fails to re-localize the object, revealing the model's lack of reasoning capacity under complete occlusion. Notably, once the occlusion resolves, the tracker successfully re-acquires the correct object at frame \emph{\#238}, suggesting that the learned representations remain discriminative when the object reappears, but an explicit occlusion-handling mechanism is needed to bridge the gap during fully occluded intervals.

\begin{sidewaystable}
\centering
\caption{PR, NPR, and SR of ESMTrack with different contribution of Cross Modality Loss}
\begin{tabular}{c c c c c c c c}
 \toprule
  $\beta$ & 0 & 0.2 & 0.4 & 0.5&0.6 & 0.8 & 1\\
  \midrule
  GTOT  & 75.2/60.7 &78.4/63.9& 78.0/63.8 & 78.9/64.3&74.2/60.7&74.5/61.0&$\bm{79.3}$/$\bm{65.1}$ \\
  RGBT234 & 66.9/47.0 &68.1/47.9& 67.0/46.8 &67.3/47.9&66.7/46.8 &66.0/46.5& $\bm{69.1}/\bm{48.5}$\\
  LasHeR & 43.8/39.7/32.9 &46.4/41.8/35.0&43.2/38.9/33.5 & 46.1/41.5/35.0 &43.4/39.2/33.3 &43.9/39.4/34.4& $\bm{47.0}$/$\bm{42.5}$/$\bm{35.1}$\\
 \bottomrule
\end{tabular}
\label{beta}
\vspace{12pt}
\centering
\caption{PR, NPR, and SR of ESMTrack with different contribution of Grounding Triplet Loss}
\begin{tabular}{c c c c c c c c}
 \toprule
  $\lambda$ & 0 & 0.2 & 0.4 & 0.5 &0.6 & 0.8 & 1\\
  \midrule
  GTOT  & 72.7/60.8 &76.6/63.7& $\bm{81.5}$/$\bm{67.7}$ &75.2/62.6 &74.6/61.9&69.2/58.2&79.3/65.1 \\
  RGBT234 & 66.8/46.5 &68.0/47.3&65.4/45.2&67.8/47.0&67.9/47.2 &62.6/43.8& $\bm{69.1}/\bm{48.5}$\\
  LasHeR &44.9/40.2/34.5 &46.0/41.3/$\bm{35.1}$& 45.4/40.4/34.2 & 46.3/41.3/34.9&45.8/40.8/34.7 &42.9/38.8/33.0& $\bm{47.0}/\bm{42.5}/\bm{35.1}$\\
 \bottomrule
\end{tabular}
\label{lambda}
\vspace{12pt}
\centering
\caption{PR, NPR, and SR of ESMTrack with different contribution of temporal triplet loss}
\begin{tabular}{c c c c c c c c}
 \toprule
  $\gamma$ & 0 & 0.2 & 0.4 & 0.5 &0.6 & 0.8 & 1\\
  \midrule
  GTOT  & 78.0/63.5 &$\bm{80.5}$/64.9 &73.7/60.0 & 72.8/59.3&79.8/65.0&72.8/59.5&79.3/$\bm{65.1}$ \\
  RGBT234 & 68.7/47.4 &64.9/45.2&66.5/46.4&66.3/46.3& 68.5/47.6 &66.5/46.5& $\bm{69.1}/\bm{48.5}$\\
  LasHeR & 46.3/42.1/$\bm{36.0}$ &44.5/40.4/34.3& 43.6/39.1/32.5&44.5/39.8/33.3 &46.3/41.4/33.7 &44.0/39.6/32.6& $\bm{47.0}/\bm{42.5}$/35.1\\
 \bottomrule
\end{tabular}
\label{gamma}
\end{sidewaystable}
\subsection{Analysis of Hyperparameters}
We conduct an ablation study on the backward IoU threshold used to filter reliable samples for the cross-frame temporal triplet loss, with results reported in Table~\ref{backward_iou}. The threshold of 0.3 consistently achieves the best performance across all four benchmarks. A lower threshold of 0.1 introduces noisy training samples whose forward tracking trajectories are unreliable, thereby degrading the quality of the temporal triplet supervision. Notably, the threshold of 0.2 yields worse results than 0.1 on all datasets, suggesting that it falls into an unfavorable regime where the sample set is neither large enough to provide sufficient supervision nor clean enough to offer reliable anchors. As the threshold increases to 0.4, performance drops consistently, indicating that overly strict filtering excludes too many valid samples per batch and weakens the triplet loss signal. These results demonstrate that a moderate threshold of 0.3 strikes the optimal balance between sample quality and quantity, and we adopt this value in all subsequent experiments.

We ablate the weight $\beta$ of the cross-modality loss while keeping other loss weights fixed. As shown in Table~\ref{beta}, removing the cross-modality loss ($\beta$=0) leads to a clear performance drop across all datasets, confirming its role in aligning RGB and infrared features. 
However, the performance is non-monotonic with respect to $\beta$: intermediate values such as $\beta$ = 0.6 and $\beta$ = 0.8 fall below or match the $\beta$ = 0 baseline on GTOT (PR: 74.2\%, SR: 60.7\% and PR: 74.5\%, SR: 61.0\% vs. PR: 75.2\%, SR: 60.7\%), suggesting that an insufficiently weighted cross-modality signal creates conflicting gradients without providing adequate alignment benefit. Performance recovers at $\beta$ = 1.0, which achieves the best results on all three datasets (GTOT: PR: 79.3\%, SR: 65.1\%; RGBT234: PR: 69.1\%, SR: 48.5\%; LasHeR: PR: 47.0\%, NPR: 42.5\%, SR: 35.1\%), indicating that full-weight cross-modality supervision provides the clearest gradient direction for modality alignment. We therefore set $\beta$ = 1.0 in all experiments.

Table~\ref{lambda} reports performance under varying $\lambda$. Without the grounding triplet loss ($\lambda$ = 0), the model achieves lower scores on all benchmarks, confirming the benefit of spatial discrimination supervision on grounding frames. The response to $\lambda$ is notably non-monotonic: $\lambda$ = 0.4 achieves the highest performance on GTOT (PR: 81.5\%, SR: 67.7\%) among all settings, but simultaneously degrades RGBT234 to (PR: 65.4\%, SR: 45.2\%), well below the $\lambda$ = 0 baseline. A more severe drop occurs at $\lambda$ = 0.8, where both GTOT and RGBT234 fall substantially (GTOT: PR: 69.2\%, SR: 58.2\%; RGBT234: PR: 62.6\%, SR: 43.8\%), suggesting that near-full weighting disrupts the balance between the main tracking objective and the auxiliary triplet signal. At $\lambda$ = 1.0, the model achieves the best aggregate performance across all three datasets, indicating that dataset-level optimization stabilizes at full weight. We therefore set $\lambda$ = 1.0 as the final configuration.

The ablation of $\gamma$ reveals a distinct pattern from the other two losses, as shown in Table~\ref{gamma}. At $\gamma$ = 0 (no temporal triplet loss), the model already achieves competitive results (RGBT234: PR: 68.7\%, SR: 47.4\%; LasHeR: PR: 46.3\%, NPR: 42.1\%, SR: 36.0\%), reflecting the contribution of grounding-only training. Notably, intermediate values of $\gamma$ from 0.2 to 0.8 frequently perform worse than $\gamma$ = 0 on RGBT234 and LasHeR—for instance, $\gamma$ = 0.2 reduces RGBT234 to (PR: 64.9\%, SR: 45.2\%). This indicates that weakly weighted search triplet signals introduce gradient noise without providing sufficient discriminative benefit, since search-frame anchors are derived from predictions rather than ground-truth annotations and are therefore less reliable than grounding anchors. Only at $\gamma$ = 1.0 does the temporal triplet loss consistently improve over all baselines, achieving the best results on all datasets. We conclude that the temporal triplet loss requires full weighting to overcome the inherent noise in search-frame anchor estimation, and thus set $\gamma$ = 1.0.
\section{Conclusion}
In this paper, we present ESMTrack, an end-to-end self-supervised RGB-T tracking framework that requires no bounding box annotations on search frames or offline pseudo-label generation. To eliminate reliance on noisy pseudo-labels and enable end-to-end optimization, we design two complementary triplet losses: a grounding triplet loss that leverages ground-truth annotations on the initial frame to provide reliable spatial supervision, and a cross-frame temporal triplet loss that maintains semantic coherence across frames, with unreliable samples filtered by backward tracking IoU. For cross-modal interaction, the Adaptive Modal Discrimination (AMD) module dynamically evaluates each modality's contribution and suppresses redundant cross-modal features, reducing shortcut dependencies and enhancing robustness to modality degradation. Extensive experiments on multiple RGB-T tracking benchmarks confirm state-of-the-art performance and validate the effectiveness of each proposed component. In future work, we plan to explore more principled modality reliability estimation, such as gradient-based importance scores or causal attribution methods, to further improve cross-modal learning under challenging degradation scenarios.
\section*{Declarations}
\begin{itemize}
\item Funding information: This work was supported by the National Natural Science Foundation of China (Grant No. 62172417) and the China Scholarship Council (Grant No. 202506420043).
\item Data availability:
The datasets used in this study are publicly available benchmarks: GTOT, RGBT234, RGBT210, LasHeR and VTUAV. The code will be made available upon acceptance of the manuscript.
\end{itemize}

\end{document}